\pdfoutput=1
\documentclass{toc}
\usepackage{xkeyval,textcomp,caption}
\makeatletter
\define@key{pubdet}{xc}[true]{\def\KV@pubdet@xc{#1}}
\makeatother
\setkeys{pubdet}{xc=true}
\tocdetails{%
  title = {Pythagorean trees for minimal reconstruction of the natural branching structures},
  number_of_pages = {NN}, 
  number_of_bibitems = {NN},
  number_of_figures = {NN},
  conference_version = {XXXXNN}, 
  author = {Dymitr Ruta, Corrado Mio and Ernesto Damiani},
  authorlist = {Dymitr Ruta, Corrado Mio, Ernesto Damiani},
  keywords = {Pythagorean tree, fractal, tree branching structure, da Vinci rule, convolutional neural network, classification}
}   

\tocdetails{%
}   

\usepackage{graphicx,amsmath,mathtools,amssymb,subfigure,listings,makecell,gensymb,colortbl,wrapfig,lipsum}

\begin{document}

\begin{frontmatter}


\title{Leonardo vindicated: Pythagorean trees for minimal reconstruction of the natural branching structures}  


\author[ruta]{Dymitr Ruta}
\author[mio]{Corrado Mio}
\author[damiani]{Ernesto Damiani}

\begin{dedication}
To my daughter Natalie
\end{dedication}

\begin{abstract}
Trees continue to fascinate with their natural beauty and as engineering masterpieces optimal with respect to several independent criteria. Pythagorean tree is a well-known fractal design that realistically mimics the natural tree branching structures. We study various types of Pythagorean-like fractal trees with different shapes of the base, branching angles and relaxed scales in an attempt to identify and explain which variants are the closest match to the branching structures commonly observed in the natural world. Pursuing simultaneously the realism and minimalism of the fractal tree model, we have developed a flexibly parameterised and fast algorithm to grow and visually examine deep Pythagorean-inspired fractal trees with the capability to orderly over- or underestimate the Leonardo da Vinci's tree branching rule as well as control various imbalances and branching angles. We tested the realism of the generated fractal tree images by means of the classification accuracy of detecting natural tree with the transfer-trained deep Convolutional Neural Networks (CNNs). Having empirically established the parameters of the fractal trees that maximize the CNN's \textit{natural tree} class classification accuracy we have translated them back to the scales and angles of branches and came to the interesting conclusions that support the da Vinci branching rule and golden ratio based scaling for both the shape of the branch and imbalance between the child branches, and claim the flexibly parameterized fractal trees can be used to generate artificial examples to train robust detectors of different species of trees.  
\end{abstract}

\end{frontmatter}

\section{Introduction}
There is no doubt natural trees are as beautiful and inspiring as they are ingeniously optimal from the engineering point of view \cite{Halle1978}-\cite{Deussen2005}, \cite{Minamino2014}-\cite{Grigoriev2022}. Ensuring the most efficient water and nutrients transport from roots to leaves up to a 100m, optimally resisting fractures from gusty winds, efficiently balancing photosynthesis with transpiration while successfully branching out to maximize the access to the space and light are just some of the tree's ingenious mechanisms that make it survive and thrive for up to thousands of years despite nature or human inflicted troubles.

Many algorithmic generative tree models have been proposed \cite{Sakaguchi1999}-\cite{Quigley2018}, some of which successfully approached photo-realism quality required for the fast expanding computer graphics and animation industry. An emerging observation from these attempts is that while being so complex, the tree appears to be also beautifully simple, with its branching structure following similar recursive generative patterns that inspired attempts to model them by fractals \cite{Prusinkiewicz1989}-\cite{Flake1998}, \cite{Bosman1957}-\cite{Grigoriev2022}. From the minimalist point of view, reconstructing the trees by fractals is the most appealing, since it offers the shortest and most essential description of how to generate the naturally looking tree structure and for that reason it is also more likely to offer explanations for the fundamental engineering principles guiding the natural tree creation, which we are keen to explore. 

What gives the natural tree its signature look is the branching structure eventually leading to the green leaves. Although in general we observe examples of trees with junctions splitting out into more than two branches, the dominant observation is, also reinforced by human imposed selective breeding, that of the apical dominance of the strongest stem (trunk), which typically leads to binary branch attachments or forking that leaves a pair of branches after the junction \cite{Halle1978}. 

There have been many attempts to describe and explain the branching structure of the natural tree \cite{Minamino2014}-\cite{Grigoriev2022}. The first dating back over 500 years ago is attributed to the Renaissance painter and polymath Leonardo da Vinci, who proposed the rule of intersection area preservation when passing through any junction of the tree. In the original Italian, the eighth rule Leonardo wrote in his notebook on drawing trees reads \emph{"Ogni biforcazione di rami insieme giunta ricompone la grossezza del ramo che con essa si congiunge"}, i.e. “all the branches of a tree at every stage of its height when put together are equal in thickness to the trunk they branch from”. This \emph{da Vinci rule} implies that the total sum of intersections of all the branches at the terminal level is equal to the intersection of the main trunk. In other words, if a tree's branches were folded upward and squeezed together, the tree would look like one big trunk with the same thickness from top to bottom. Such model gained the support of the \textit{pipe model} \cite{Shinozaki1964}, that considers the tree a collection of vascular vessels connecting roots to the leaves, as well elastic similarity model \cite{Mcmahon1976},\cite{West2000}, which assumes fixed rate of branches weight-imposed deflection per unit of their length, later proven to result with the optimal (even) distribution of fracture risk implying globally minimized probability of fracture if the da Vinci rule truly holds \cite{Eloy2011}.  
However simple and beautiful the da Vinci rule may seem, recent reviews analyzing experimental work conclude that it does not hold in general conditions \cite{Lehnebach2018}. A pipe model has been partially dismissed due to the fact that the percentage of the tree cross-section responsible for the vascular transport appears to be only around 5\% and its active \textit{biomass} is distributed just under the surface. Earlier experimental work in \cite{Sone2005}, hinted at the evidence that the total intersection area of the children branches grows faster than the parent's intersection. 

Other researchers relied on simulation, examining da Vinci's rule at the light of bio-mechanical models. A study by R. Minamino and M. Tateno \cite{minamino2014tree} computed the ratio between the cross-sectional area of the branches to the cross-sectional area of the trunk at the branching point, showing that bio-mechanical models agree with da Vinci's rule when the branching angles of daughter branches and the weights of lateral daughter branches are small; however, the models deviate from da Vinci's rule as the weights and/or the branching angles of lateral branches increase. 

Grigoriev et al. \cite{Grigoriev2022} disputed that the length of the branch is not sufficiently accounted for in the branching models, while pointing at the evidence that longer than expected branches are also thinner than what would be expected from the da Vinci rule. Based on the Fourier intensity plot analysis of the tree images he further claims that the tree branching structures follow the behavior of the logarithmic fractal structures that preserve the rule of the total surface area when crossing through the junctions instead of da Vinci's intersection area \cite{Grigoriev2022}. This major adjustment proposition seems to explain much broader tree species including very different branching structures of the slender birch and the bulky branches of the oak trees. This theory received some backing of some micro-biologist community on the grounds that the alive part of the tree truly resides in its outer layer just below the surface, hence it is the branches' surface area instead of intersection what appears to best guide the branching structures of the natural trees.

Recently, S. Sopp and R. Valbuena \cite{sopp2023}
proposed a metabolic approach to organism scaling and described a model based on \emph{allometry}. The idea is that real trees do not preserve branches shape or size, but rely on an external metric, hydraulic resistance, adjusting the rate of conduit widening and coalescence to preserve it. According to the authors, hydraulic resistance preservation optimizes the carbon employed to develop the tree vascular system. From this principle, Sopp and Valbuena derived an elegant inverse relationship between the widening of conduits from the tree top to base and the tapering of branch volume from the base to the top. Their relationship contradicts - or better, improves on - the da Vinci rule, providing results closer to experimental evidence on tree growth. Still, even Sopp and Valbuena's inverse relation achieves its elegance at the expense of some simplification, as their underlying growth model limits conduit coalescence to the distal part of the conductive system. In the future, more complex metabolic models of tree growth may emerge that improve accordance with experimental data by taking into account tissue micro-structures or even plant genetics. Still, the increase in complexity is likely to make such models cumbersome.

In this paper we take a different approach: rather than using advanced knowledge on plant metabolism to derive a rule that matches available experimental data, we improve on da Vinci's rule by providing an agile mathematical framework for the construction of branching structures; then we use Machine Learning to assess their closeness to natural ones. 
We aim to show that the intuition behind da Vinci's rule still holds: mathematically minimal reconstructions can provide pictorial representation of natural branching structures that provide the visual experience of harmony and beauty of the natural tree shape.

With all the evidence at our disposal, Bosmans's Pythagorean tree fractals \cite{Bosman1957} appear to be the best starting point of our investigation focused on the simplest and most realistic fractal reconstruction of the natural tree.

The remainder of the paper is organized as follows. A general description of Pythagorean trees model for generating fractal trees with visual examples is provided in Section \ref{sec:pytree}. Their parametric generalization accounting for da Vinci's rule and variable branching angle, along with an efficient recursive software implementation are introduced in Section \ref{sec:ftree}. Section \ref{sec:cnn_tree_clasf} covers detailed experiments with several Convolutional Neural Networks (CNN) transfer-trained to detect natural trees in images and deployed to classify various families of generalized fractal trees generated from our parametric model and attempts to identify which parametric combinations produce the most naturally looking fractal trees. Finally brief conclusions and plans for further research are drawn in Section \ref{sec:Conclusions}.

\section{Pythagorean tree}\label{sec:pytree}
\begin{figure}[hbt!]
    \centering
    \subfigure[D-5 Pythagoras tree]{\includegraphics[width=0.49\hsize]{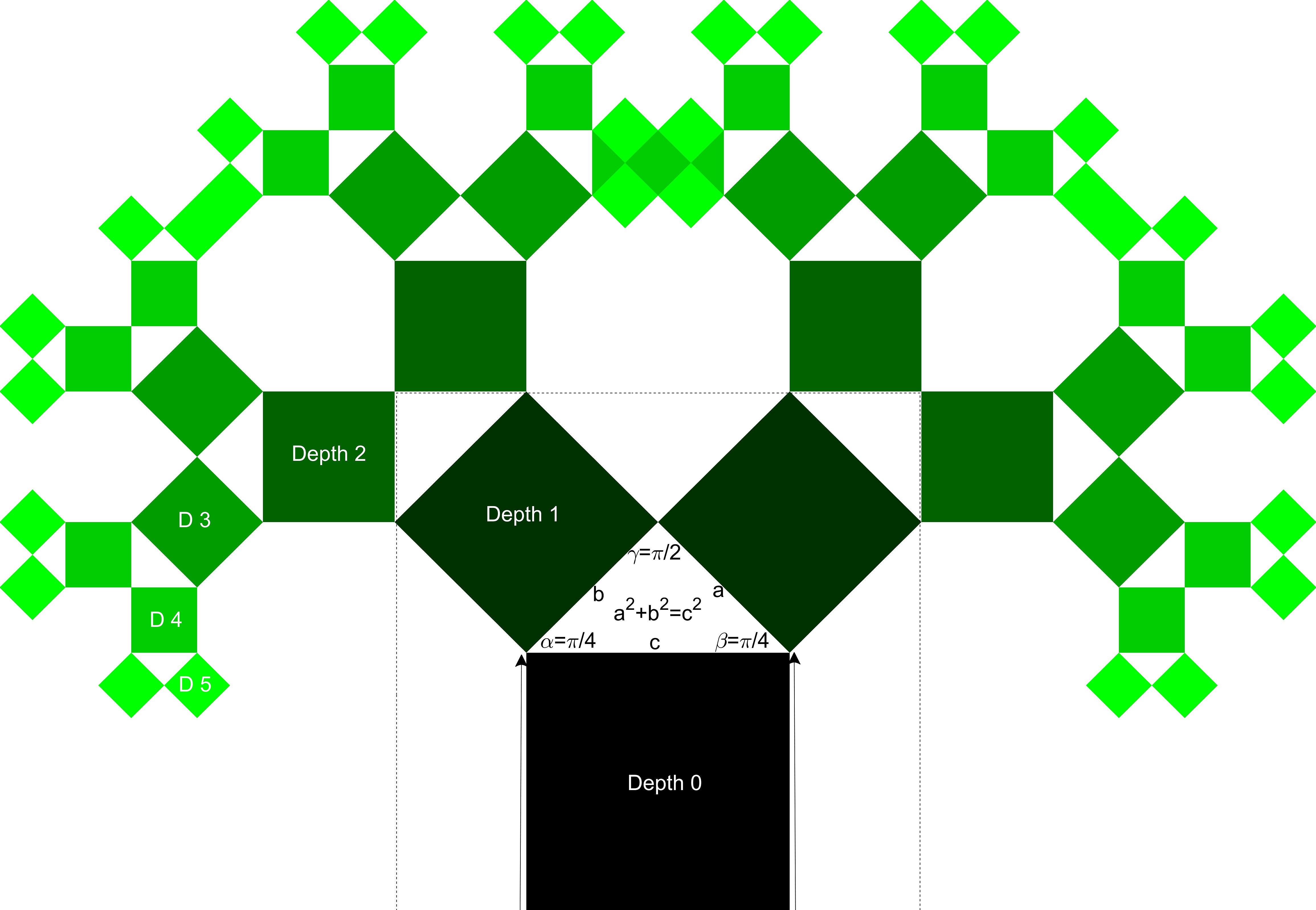}\label{fig:ptree_demo_a}}
    \subfigure[D-25 Pythagoras tree]{\includegraphics[width=0.49\hsize]{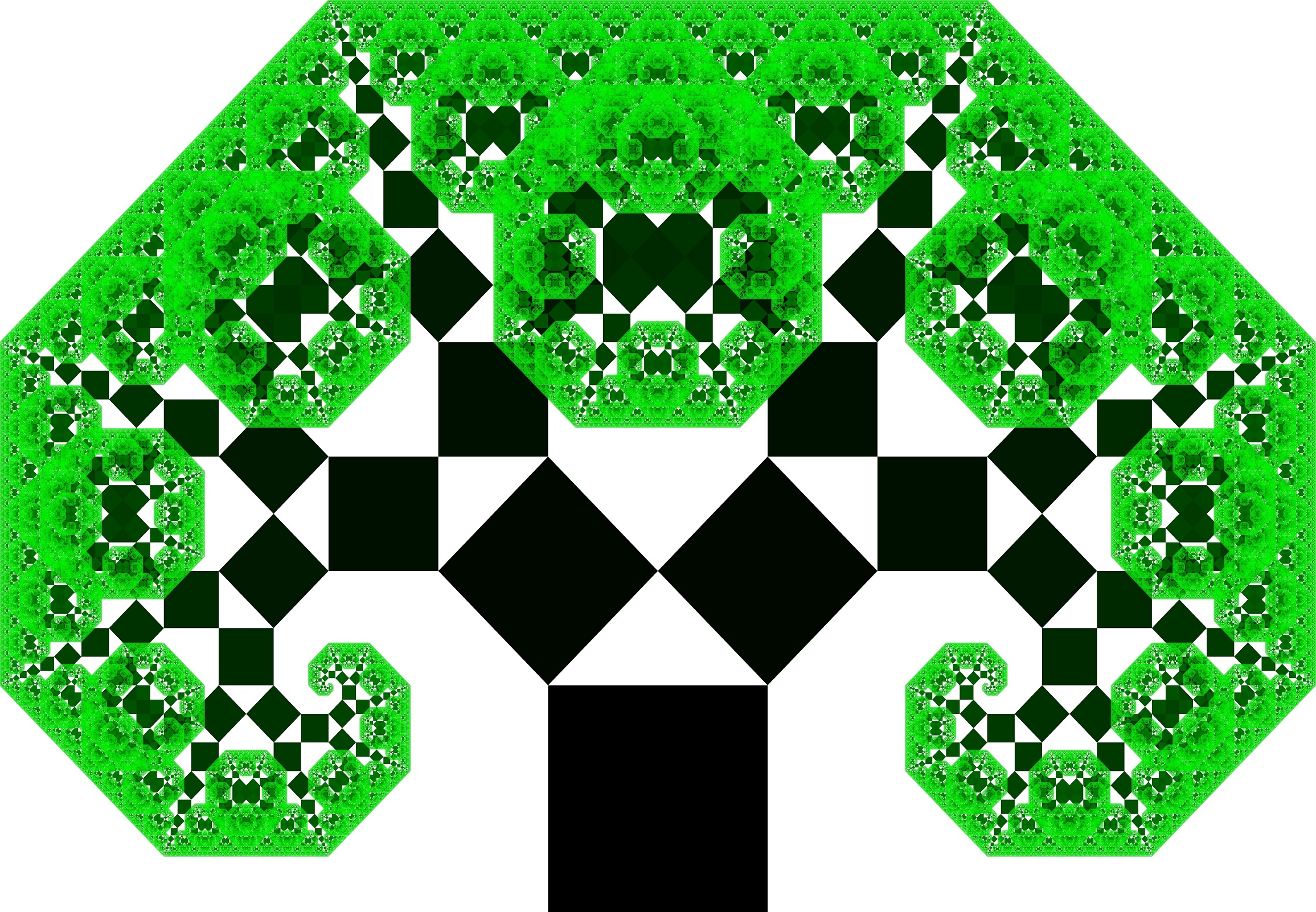}\label{fig:ptree_demo_b}}
    \caption{Pythagorean trees with depth of 5 and 25. At each recursive step a pair of child squares is generated by scaling the parent square down with a scale $s=\sqrt(1/2)$, shifting them up to the parent upper corners and rotating them by $\pm=\pi/4$ angles} 
    \label{fig:ptree_demo}
\end{figure}

Pythagorean tree is a plane fractal design constructed from squares resting on the sides of the right triangle. It was first described in 1942 by Albert E. Bosman  who coined  the name to highlight the link to the famous ancient Greek mathematician.  The base fractal configuration of squares enclosing the right triangle is also traditionally used to illustrate the Pythagorean Theorem \cite{Bosman1957}. It is perhaps worth noting that Elisha Scott Loomis' classic “The Pythagorean Proposition” \cite{loomis1968pythagorean}, which contains 371 proofs of the Pythagorean Theorem, includes one by Leonardo da Vinci.
In its simplest form with isosceles right triangle, construction of symmetrical Pythagorean tree recursively takes the base square and puts its $\sqrt(1/2)$-scaled down and $\pm\pi/4$-rotated versions around the parent upper corners as shown in an instructive illustration of 5-levels deep Pythagorean tree in Figure \ref{fig:ptree_demo_a}, followed with deep expanded tree up to $25$ recursive steps in Figure \ref{fig:ptree_demo_b}. 

\begin{figure*}[hbt!]
    \centering
    \includegraphics[width=1\hsize]{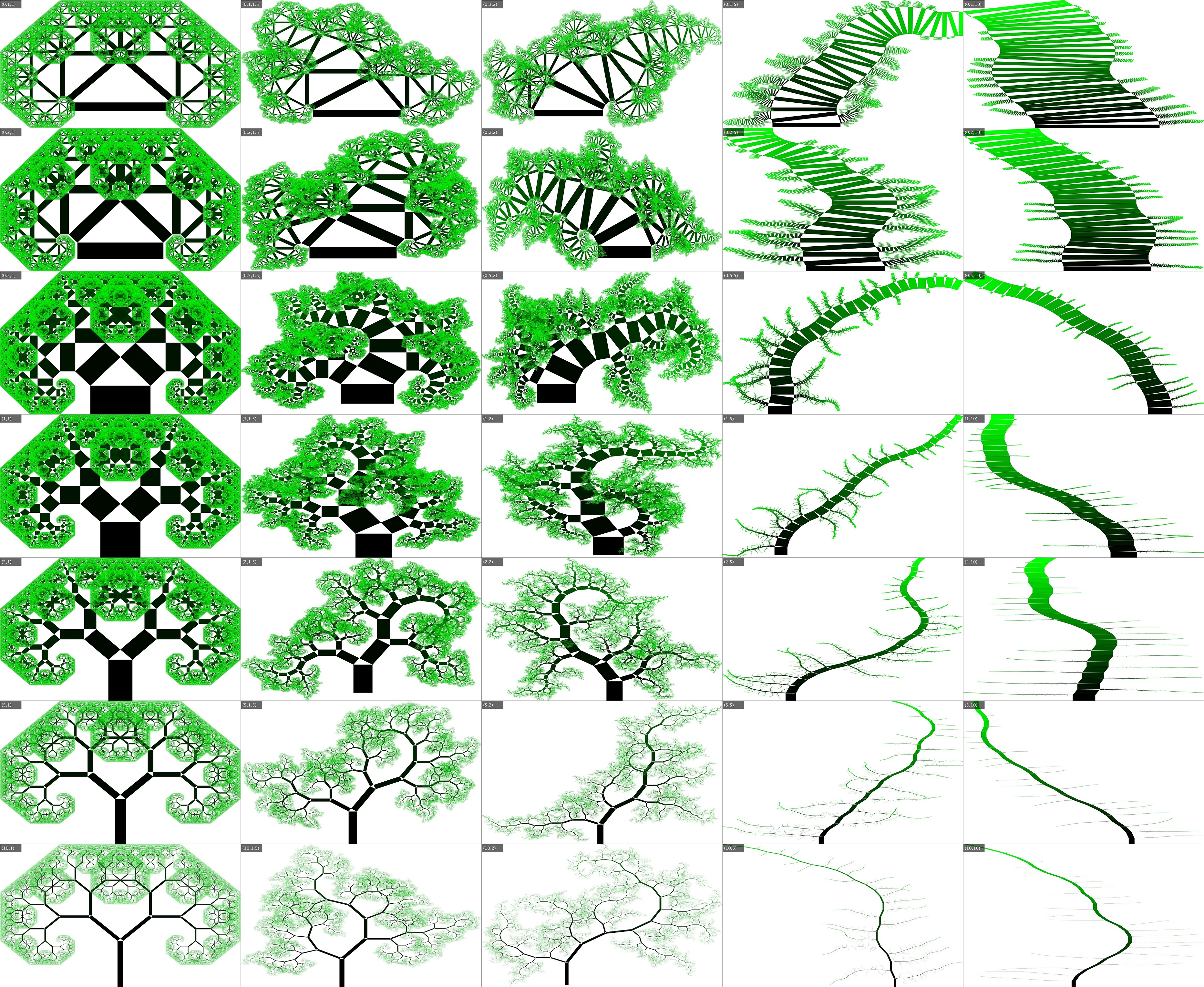}
    \caption{Pythagorean trees grown recursively (26 depths) with all the unique combination of elongations $e$ (rows) and branch imbalance $b$ (columns) $(e,b): e=[.1,.2,.5,1,2,5,10], \; b=[1,1.5,2,5,10]$} 
    \label{fig:ptree_shape_param}
\end{figure*}

For clarity, we assume the root at depth $0$ and introduce a simple tree-mimicking color map linearly scaling the colors from full black to full green along the recursion depth. An immediate observation is that the model with squares and isosceles right triangle appears too simple and always produces the same shape of the tree, that although looks like a tree is very far from realistic appearance of the natural tree. We therefore introduced two levels of parameterization and a simple randomisation element to the tree growth algorithm. Firstly, we have allowed for the base element to be a rectangle with an elongation shape parameter $e$ expressing the ratio of its height to width and the branching imbalance parameter $b$ expressing the ratio of the larger to smaller sides of the right branching triangle that models the diameters of the forked branches. Moreover, at each step we introduce a random flip of the branching triangle to avoid unrealistic bending of the branched out children always in the same direction and by the same rotation angle. $35$ Pythagorean trees have been generated this way for the grid of parameters $(e,b)$: $e=[0.1,0.2,0.5,1,2,5,10]$ and $b=[1,1.5,2,5,10]$, corresponding to rows and columns of the grid with 26-levels deep Pythagorean trees shown in Figure \ref{fig:ptree_shape_param}.

Some of the trees appear much more realistic than others. The closest to reality from this set seem the trees with elongated base rectangles ($e=2-10$) and slightly imbalanced branches' diameters ($1<b\leq 2$). Interestingly, the trees with $e<1$ completely lose the tree look, starting to resemble the rooting or "hairy" stem structures when accompanied with large branch imbalance ($b\geq5$) and maintain this resemblance also for elongated ($e>1$) branches. It appears that randomized branching triangle flip contributed mostly to the more realistic looks of the fractal trees, although from the engineering point of view when the reduction of the width of the main branch is small ($b>2$) we may see a too large mass of the tree tilting sideways that in reality would collapse under its weight pulled down by gravity.   

It is important to mention that due to the right-angled branching triangle the sums of the squared scales of the children branches always add up to 1, hence Pythagorean trees meet both the da Vinci intersection preservation and Grigoriev's surface area preservation rules at every junction.

\section{Generalised fractal trees}\label{sec:ftree}
In an attempt to further encourage the more realistic appearance of the Pythagorean trees we have decided to allow for deviation from the Pythagoras rule of the branching triangle that would allow any triangle shape, while maintaining random flip in the recursive growth. To orderly parameterize the freed shape of the branching triangle we kept the branch imbalance parameter $b$ unchanged from the Pythagorean trees but added the variable $\alpha$ of the triangle's angle opposite its resting base $a$. Moreover, we also decided to put the da Vinci's rule to test by actively controlling the fraction by which the total intersection of the branches reduces or exceeds the intersection of the parent. To achieve this we simply multiply the scales derived from the sides of the branching triangle by a fixed scalar computed to achieve desired ratio $v$ of the branches to parent intersection. For notation simplicity we will consider a fractal tree defined by the $4$ parameters as: $T(e,b,\alpha,v)$ in the subsequent analysis.

Given the tree input parameters $(e,b,\alpha,v)$ and the unconstrained shape of the triangle resting on a rectangular parent with sides $(1,e)$ we can apply the cosine rule to the branching triangle to derive the scales for the sides of the branching triangle.
Book II of Euclid's Elements, compiled c. 300 BC from material up to two centuries older, contains a geometric theorem corresponding to the law of cosines but expressed in terms of rectangle areas. Euclid proved the result by using the Pythagorean Theorem. Using modern notation, the rule can be written as follows:
\begin{equation}
    s_r=1/\sqrt{1+b^2-2bcos(\alpha)}, \;\;\; s_l=bs_r
\end{equation}
Now given all the sides of the triangle and the same cosine rule we can find the rotation angles required to construct the child branches:
\begin{equation}
    \gamma=\arccos{\frac{s_l^2-s_r^2+1}{2s_l}}, \;\;\; \beta=\pi-\alpha-\gamma;
\end{equation}

\begin{figure}[hbt!]
    \centering
    \includegraphics[width=.98\hsize]{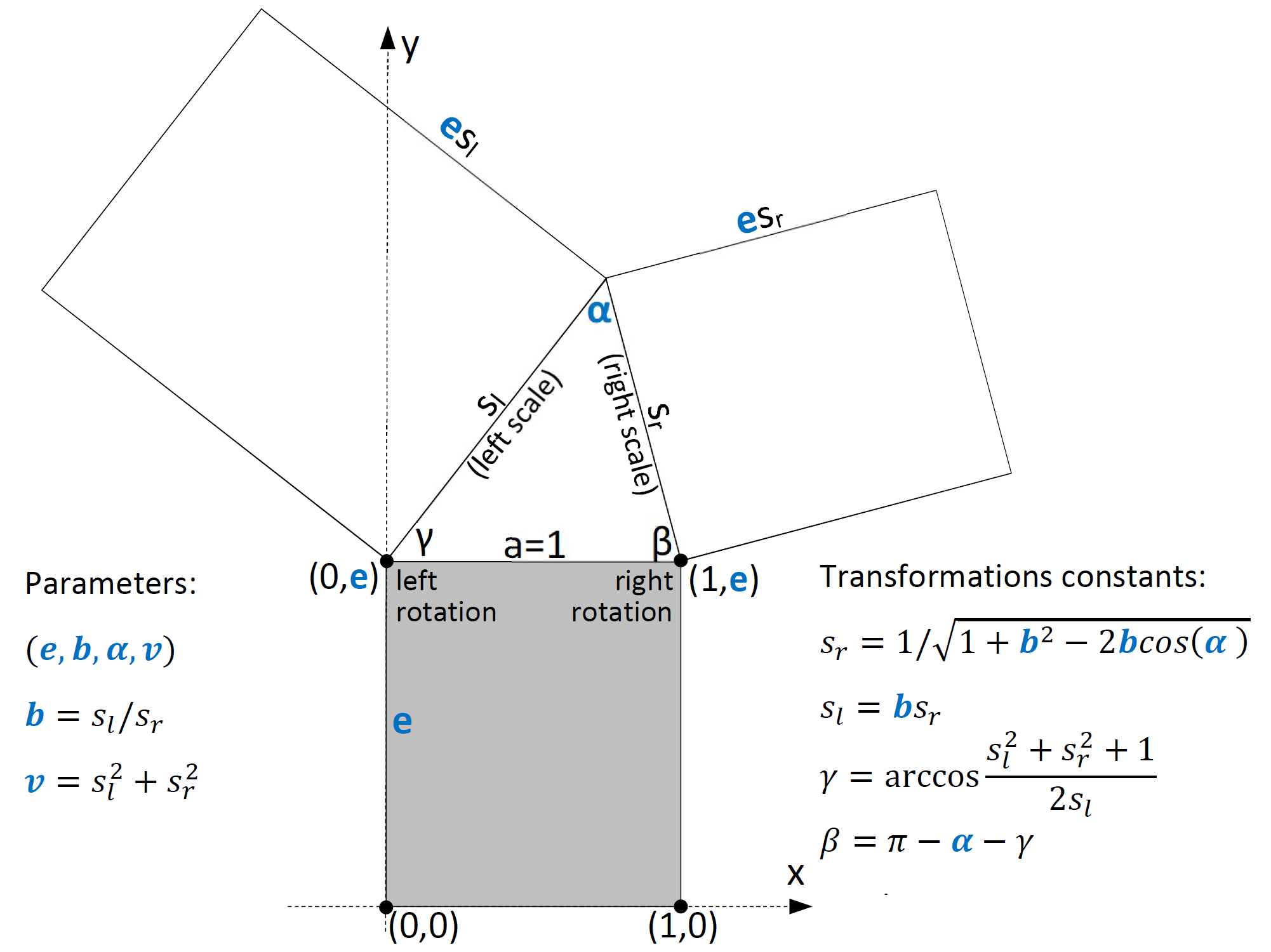}
    \caption{Illustration of the generalized fractal tree generator parameters (in blue) and derived constants used in the main recursive function}
    \label{fig:algo_params_constants}
\end{figure}

The parameters and their use to derive key fractal transformation constants required by the main recursive tree generator function are depicted in Figure \ref{fig:algo_params_constants}.

With the computed scales and rotation angles both child branches can be instantly computed from the parent simply by scaling the parent down with respect to its bottom vertices, translate (shift) them "up" by the vectors constructed from parent sides and rotating the left branch counterclockwise by an angle $\gamma$ and right branch clockwise by $\beta$. In order to enforce ($v=1$) or violate ($v\neq1$) da Vinci rule, that for scales of branches means $s_l^2+s_r^2=1$, the scales  $s_l,s_r$ have to be further multiplied by the factor $f$ computed as follows:
\begin{equation}
    f=\sqrt{\frac{v(b^2-2cos(\alpha)+1)}{b+1}}
\end{equation}
Note that since we departed from the Pythagorean constraint of branching with the right-angled triangle, enforcing da Vinci rule for obtuse angle: $\alpha>\pi/2$, would lead to additional thickening of children widths with their partial overlap, that is observed in nature, while thinning and disconnecting the branches for acute $\alpha<\pi/2$, that is rather not expected in nature. For that reason we will explore the sensitivity of the fractal tree looks to the da Vinci rule-implied children branch sizes shifted slightly towards the excess or deficiency over/under the da Vinci parameter $v=1$, for example by setting the explored grid values to $v=\{0.9,1,1.1,1.25\}$   

\subsection{Fast recursive software implementation}
The coding strategy for fractal implementation is simple: take the object, compute its perturbed, smaller, connected version(s) and recursively repeat the same upon the reduced versions until the depth limit is reached. In our case there is no difference, the recursive function tree() will be launched with the initial rectangular base and the pre-computed scaling and rotation constants for left and right child and the depth limit $d$:

\begin{lstlisting}
t=tree([0 1 1 0;0 0 e e]',sl,rotl,sr,rotr,d);
\end{lstlisting}

To further simplify and speed up the rotation process inside the function we have pre-computed rotations in a matrix form:
\begin{equation}
    rot_l=\begin{bmatrix}cos(\gamma) & sin(\gamma)\\-sin(\gamma) & cos(\gamma)\end{bmatrix}, \;
    rot_r=\begin{bmatrix}cos(\beta) & sin(\beta)\\-sin(\beta) & cos(\beta)\end{bmatrix}
\end{equation}
and they are passed as constant matrix parameters to the recursive function such that internal computations are simplified to the optimally vectorized computation of the rotated branches. Overall, given the parent trunk $t$ and the parameters, the whole process of scaling, translating and rotating that produces the children branches can be simplified to the following:

\begin{lstlisting}
function t=tree(t,sl,rotl,sr,rotr,d)
dx=t(end,:)-t(1,:);            %Fixed shift up
tl=t(1,:)+sl*(t-t(1,:))+dx;    %Left scale & shift
tr=t(2,:)+sr*(t-t(2,:))+dx;    %Right scale & shift
tl=tl(1,:)+(tl-tl(1,:))*rotl;  %Left rotate
tr=tr(2,:)+(tr-tr(2,:))*rotr'; %Right rotate
\end{lstlisting}

noting the transposed \texttt{rotr} since rotation of the right branch happens clockwise. Once the child branches are computed what remains to be done is to recursively call the same tree() function upon freshly generated children branches and/or organize the output depending on the depth we are in:

\begin{lstlisting}
if d==1
    t=[t;tl;tr];
else
    t=[t;tree(tl,sl,rotl,sr,rotr,d-1)];
    t=[t;tree(tr,sl,rotl,sr,rotr,d-1)];
end
\end{lstlisting}

To implement the random flip of the branching triangle we need to make provision to run the transformations on the swapped scales and rotations if a random flip turns opposite:  
\begin{lstlisting}
if rand>.5
    tl=t(1,:)+sr*(t-t(1,:))+dx;
    tr=t(2,:)+sl*(t-t(2,:))+dx;
    tl=tl(1,:)+(tl-tl(1,:))*rotr;
    tr=tr(2,:)+(tr-tr(2,:))*rotl';
end
\end{lstlisting}

\subsection{Visual review of the generalized fractal trees}
The optimized recursive code described above allowed to generate fractal trees in a fraction of a second up to the depth of 18. Deeper trees required essentially doubled processing time for each next depth such that 25-deep tree took around 2 minutes to generate. At depth=25 displaying the tree takes almost double the generation time since it requires plotting about 64 million coloured rectangles with 256m edges and vertices. At this depth we have generated 64 generalized fractal trees for all combinations of the four critical parameter values split into a grid of 4 values: $b=\{1,1.25,1.5,2\}$, $\alpha=\{60\degree,90\degree,120\degree,150\degree\}$, $v=\{0.9,1,1.1,1.25\}$, except of elongation that we kept fixed at $e=5$ for which we received the most realistic results with the Pythagorean trees. For simpler visual association we replace $\alpha$ with the corresponding angle between branches that is always exactly $\pi-\alpha$. 

\begin{figure}[hbt!]
    \centering
    \includegraphics[width=1\hsize]{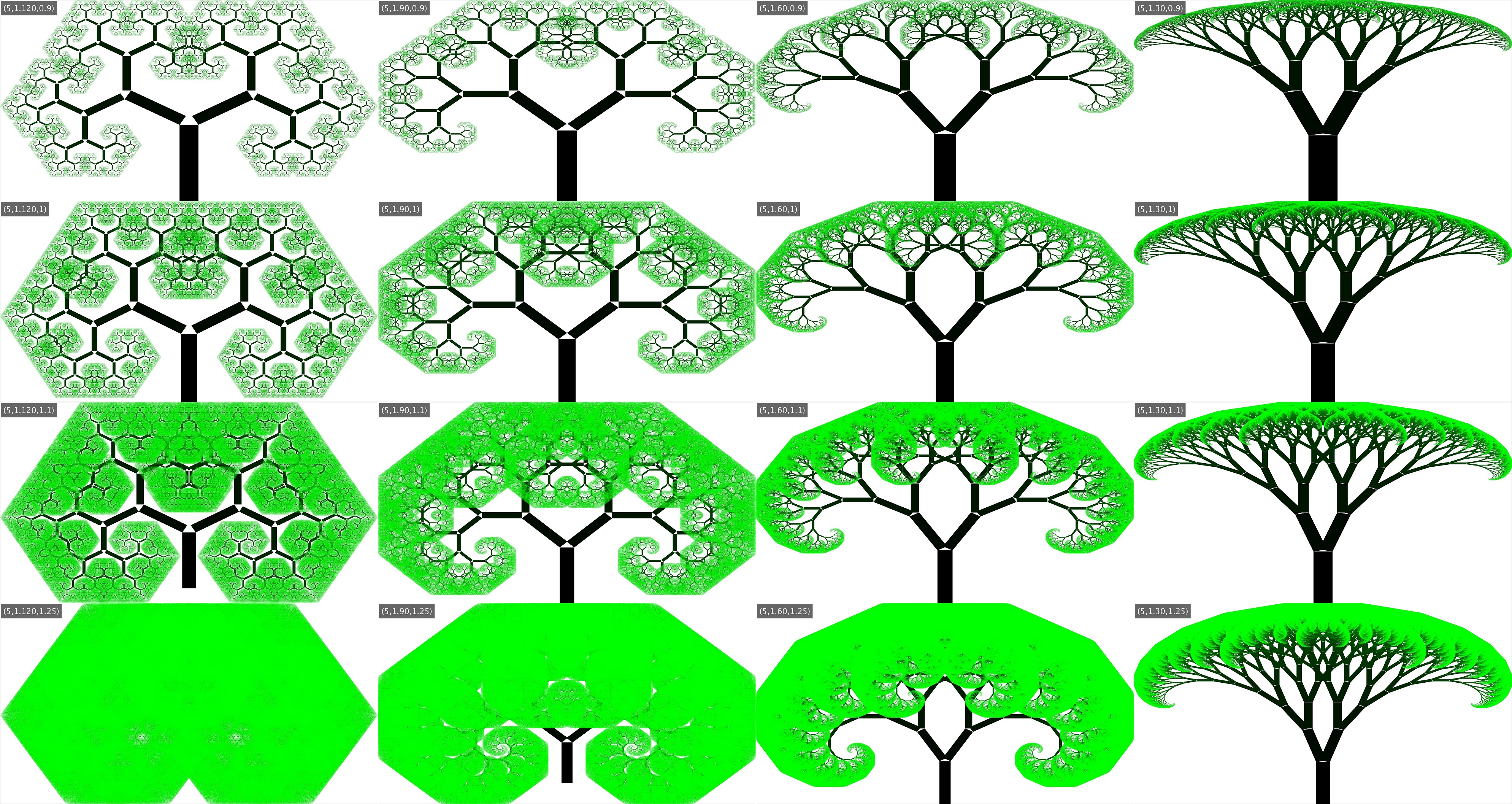}
    \caption{Generalized fractal tree for parametric combinations $T(5,1,\alpha,v)$}
    \label{fig:rtree_5_1}
\end{figure}

Figure \ref{fig:rtree_5_1} illustrates the results of the trees $T(5,1,\alpha,v)$ for the combination of parameters with $b=1$ to assess the sensitivity to branching angle $\alpha$ and the da Vinci factor $v$ for the symmetric-balanced trees. 

\begin{figure}[hbt!]
    \centering
    \includegraphics[width=1\hsize]{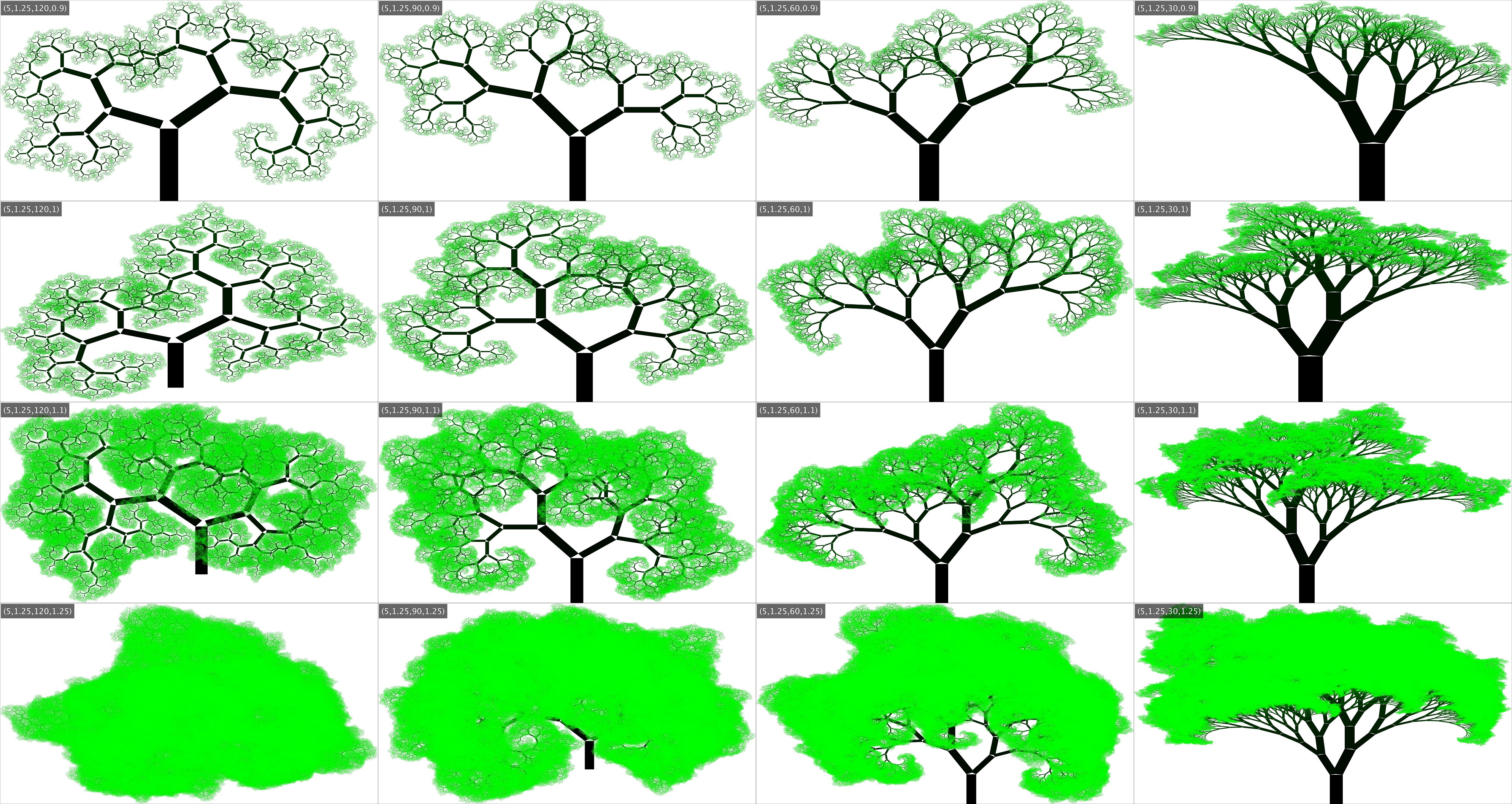}
    \caption{Generalized fractal tree for parametric combinations $T(5,1.25,\alpha,v)$} 
    \label{fig:rtree_5_125}
\end{figure}

We can observe how the traditional look of Pythagorean tree gradually evolves towards a kind of Akacia tree by narrowing branching angles $\alpha$ which translate into smaller spread between the branches, and at the same time how much more fuller and greener the tree becomes for growing da Vinci factor $v$.
For the corresponding trees generated with slight imbalance ($b=1.25$) i.e. for parametric combinations $(5,1.25,\alpha,v)$ presented in Figure \ref{fig:rtree_5_125}, the trees become much more realistic especially in the middle section with $60\degree \leq\alpha\leq 90\degree$ and $1\leq v\leq 1.1$

\begin{figure}
    \centering
    \includegraphics[width=1\hsize]{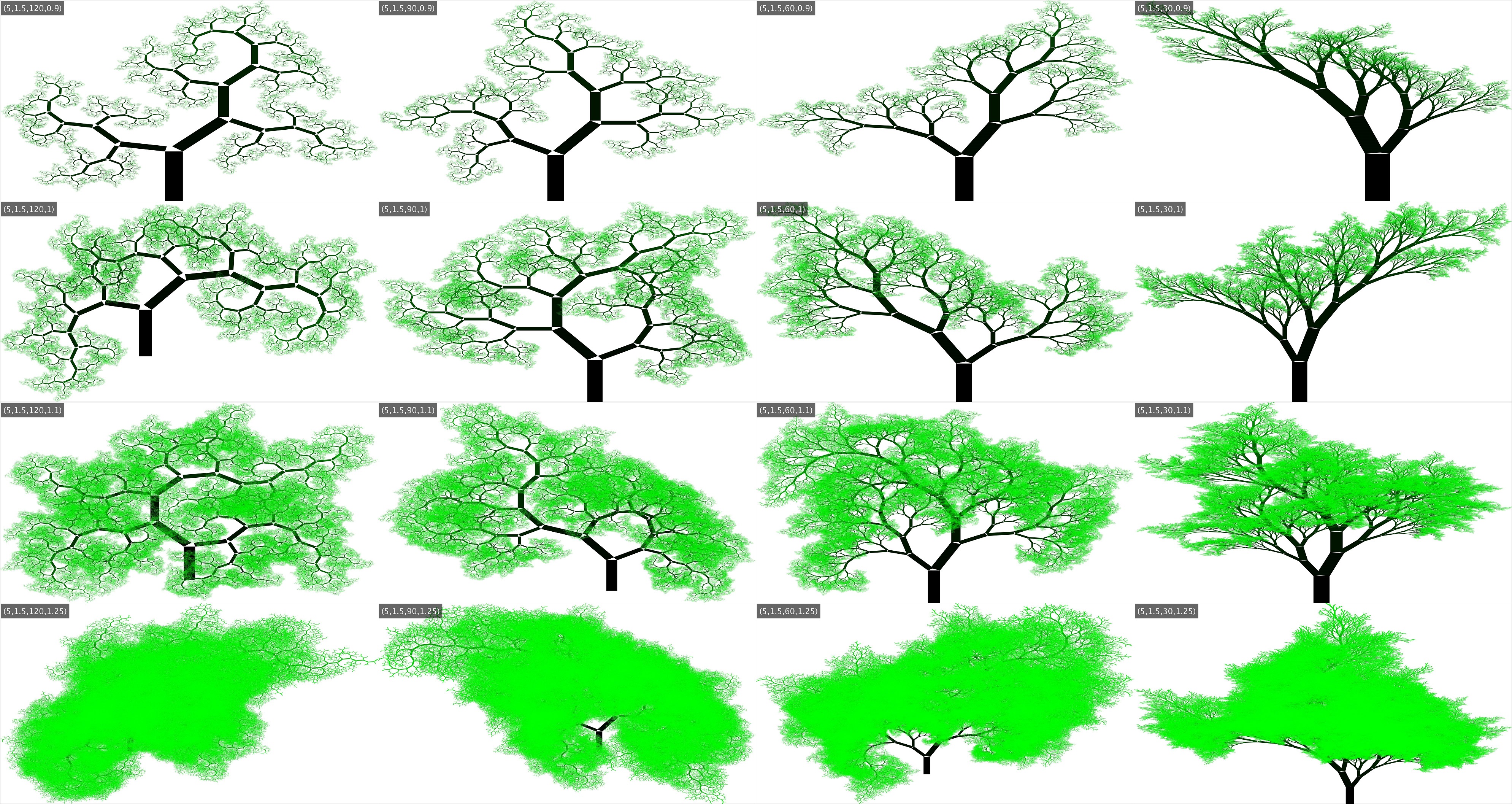}
    \caption{Generalized fractal tree for parametric combinations $T(5,1.5,\alpha,v)$} 
    \label{fig:rtree_5_15}
\end{figure}

\begin{figure}
    \centering
    \includegraphics[width=1\hsize]{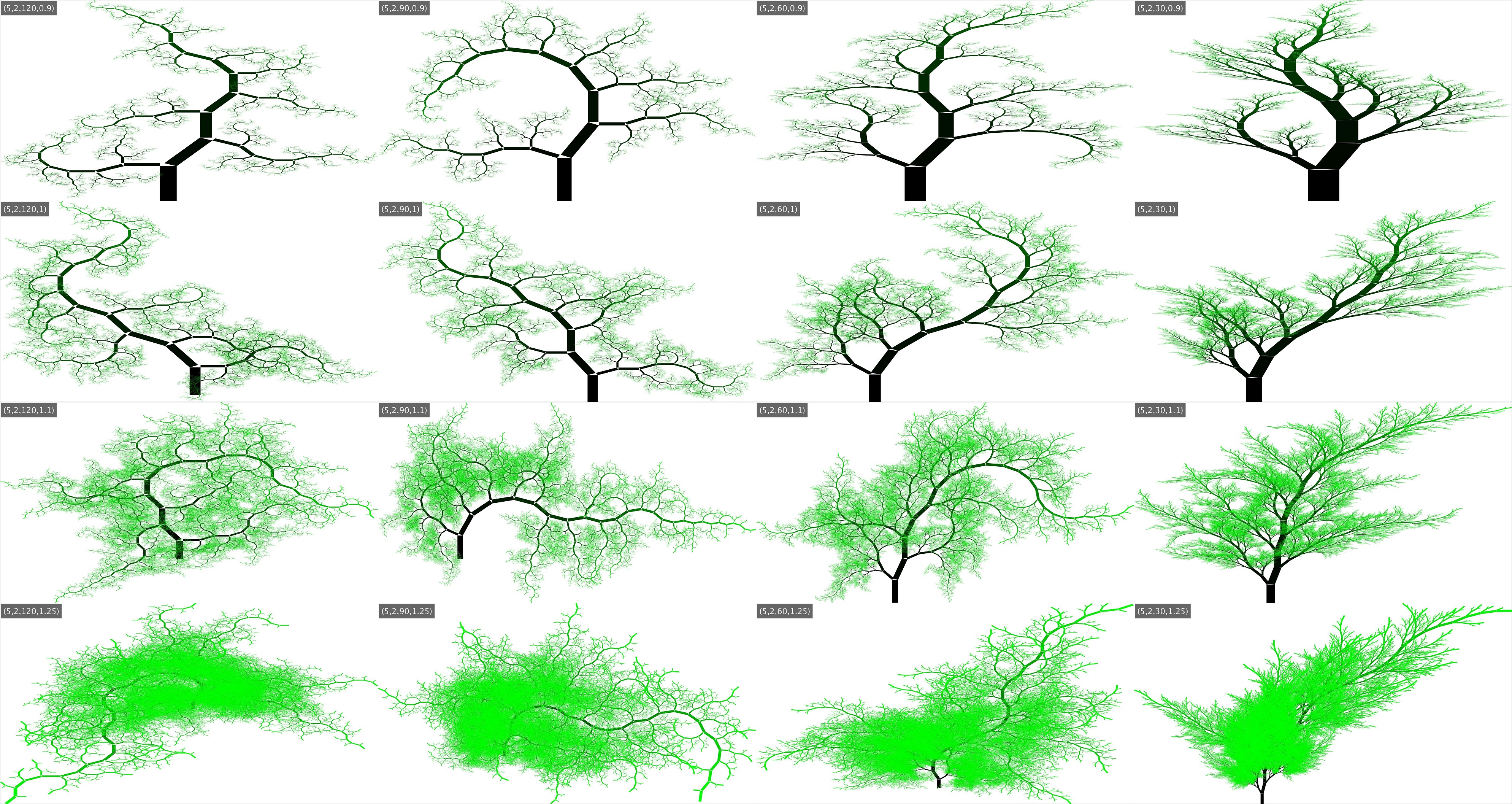}
    \caption{Generalized fractal tree for parametric combinations $T(5,2,\alpha,v)$} 
    \label{fig:rtree_5_2}
\end{figure}

Further tree plots for higher imbalance i.e. for $(5,1.5,\alpha,v)$ and $(5,2,\alpha,v)$ shown in Figures \ref{fig:rtree_5_15} and \ref{fig:rtree_5_2} confirm the same trends, although the realism sweet spot shifts slightly to the right, i.e. towards narrower branching angles, while the left and bottom sections evolve towards rooting structures. 

\section{Deep learning for fractal trees classification}\label{sec:cnn_tree_clasf}
\begin{figure}[hbt!]
    \centering
    \includegraphics[width=1\hsize]{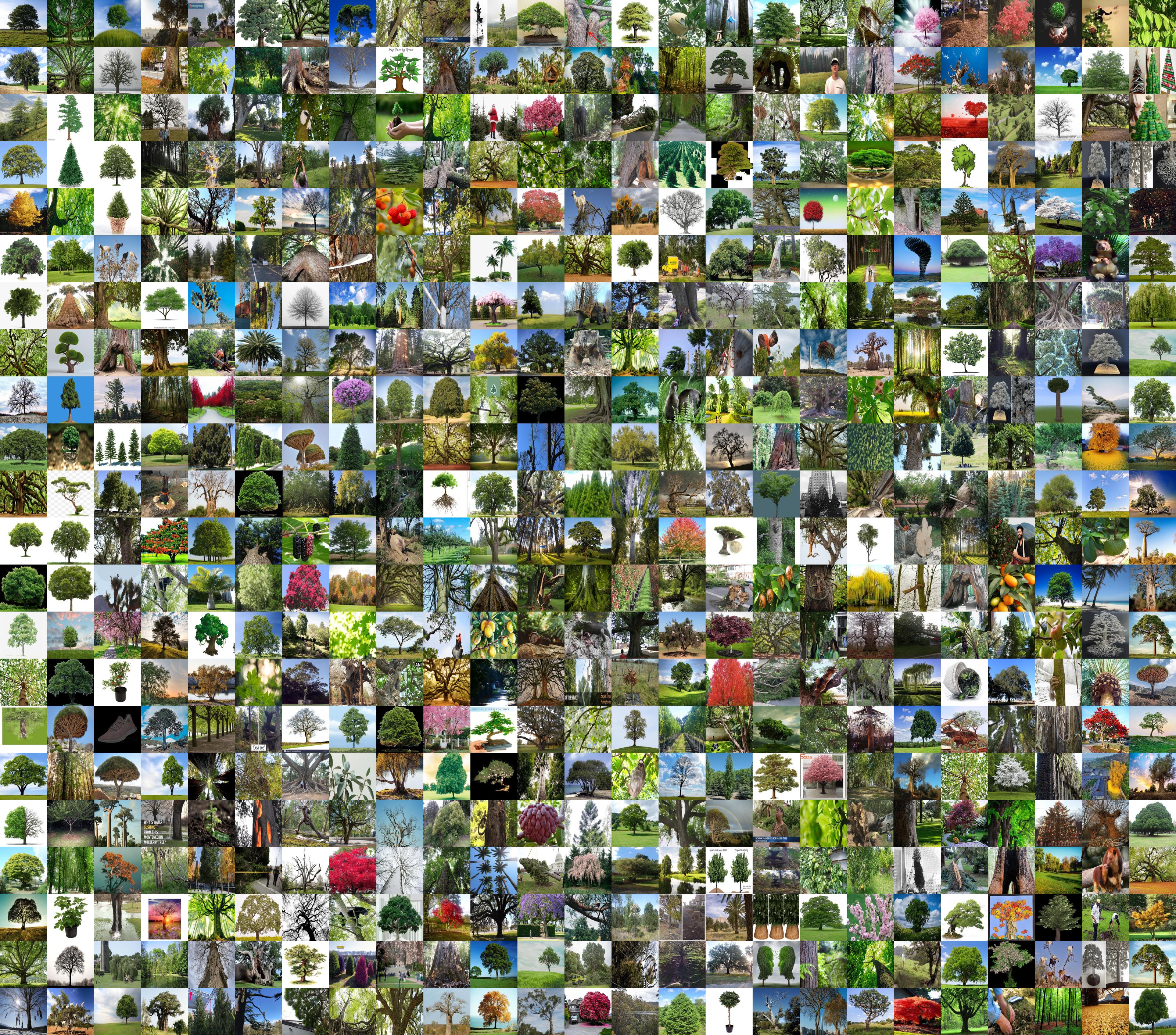}
    \caption{Mosaic of the tree class examples from updated DomainNet dataset \cite{DomainNet}} 
    \label{fig:tree_class}
\end{figure}

Overall 99 fractal trees (35 of which are Pythagorean) have been generated and visually assessed for subjective resemblance to the real trees. While the most realistic trees appear to be $T(5,1.25:1.5,60\degree:90\degree,1:1.1)$, we set off to evaluate the realism of the fractal trees by means of supervised Machine Learning (ML). Specifically, we chose the state-of-the-art deep convolutional neural networks (CNN) re-trained (transfer-trained) on real tree images along with hundreds of other objects observed in everyday life and deployed to classify the images of our fractal trees. To build the CNN we initially used the established Googlenet \cite{GoogleNet} design originally pre-trained on \textit{ImageNet} \cite{ImageNet} dataset and attempted to transfer-train it on the real part of the DomainNet \cite{DomainNet} dataset pooling together over 170k images of various objects grouped in the 345 class-categories, one of which was obviously the natural tree. Given the tree class was rather small compared to other classes we have complemented it by a number of real tree images extracted from Google image search. The tree class is representatively illustrated in Figure \ref{fig:tree_class}.

The default Googlenet architecture was updated to accommodate DomainNet's 345 classes. All images were reduced to the Googlenet accepted image input size of $224\times 224$ and the network re-trained for up to 100 epochs with batch size of 128 images and the fixed learning rate of $0.0001$, using 10\% of the data for validation and applying every-epoch-shuffling. After about 50 epochs the validation classification accuracy saturated at around 80\%, even though the training accuracy kept rising up above 90\% by the end of 100 epochs, at which stage the network was clearly over-fitted with validation accuracy falling back below 75\%. We have therefore collected the most robust network re-trained after 50 epochs and applied to classification of our fractal images, also reduced to the same compliant size of $224\times 224$. The classification results were collected in the form of probability of (support to) the \textit{tree} class, as well as up to 3 top ranked classes along with their soft probabilities all organized along descending order of the probability of \textit{tree} class as shown in Table \ref{tab:performance_results}.  

\captionsetup[table]{skip=1pt}
\begin{table}
\footnotesize
\centering
\caption{Googlenet classification results for the Pythagorean and generalized fractal tree images}
\label{tab:performance_results}
\renewcommand{\arraystretch}{0.4} {
\resizebox{\textwidth}{!}{
 \begin{tabular}{|r|l|l|r|l|r|r|r|r|r|r|r|}
\hline
\rowcolor{gray!20} type & e & b & $\alpha$ & $v$ & p(tree) & class1 & class2 & class3 & p(class1) & p(class2) & p(class3) \\ [.4ex] 
\hline
 ftree& 5.0000& 1.5000&   90& 1.1000&0.9897&          tree&          leaf&        grapes& 0.9897& 0.0053& 0.0021 \\
 ftree& 5.0000& 2.0000&   90& 1.0000&0.9890&          tree&          leaf&     palm tree& 0.9890& 0.0064& 0.0016 \\
 ftree& 5.0000& 2.0000&   60& 1.0000&0.9855&          tree&          vase&   house plant& 0.9855& 0.0050& 0.0019 \\
pytree& 5.0000& 2.0000&   90& 1.0000&0.9717&          tree&         broom&          leaf& 0.9717& 0.0120& 0.0098 \\
pytree& 2.0000& 1.5000&   90& 1.0000&0.9710&          tree&      broccoli&        grapes& 0.9710& 0.0201& 0.0063 \\
pytree& 2.0000& 2.0000&   90& 1.0000&0.9568&          tree&          vase&   house plant& 0.9568& 0.0167& 0.0085 \\
 ftree& 5.0000& 1.5000&  120& 1.0000&0.9483&          tree&     palm tree&   house plant& 0.9483& 0.0428& 0.0032 \\
 ftree& 5.0000& 1.5000&   60& 0.9000&0.9182&          tree&           map&       giraffe& 0.9182& 0.0182& 0.0119 \\
 ftree& 5.0000& 1.5000&   90& 1.0000&0.8896&          tree&      broccoli&           map& 0.8896& 0.0216& 0.0121 \\
 ftree& 5.0000& 1.5000&  150& 1.2500&0.8324&          tree&     palm tree&   house plant& 0.8324& 0.1632& 0.0013 \\
 ftree& 5.0000& 1.5000&  120& 0.9000&0.8164&          tree&     palm tree&   house plant& 0.8164& 0.1403& 0.0396 \\
 ftree& 5.0000& 1.5000&   60& 1.0000&0.8112&          tree&        grapes&      broccoli& 0.8112& 0.1476& 0.0188 \\
 ftree& 5.0000& 2.0000&   60& 0.9000&0.7926&          tree&          vase&         broom& 0.7926& 0.1661& 0.0180 \\
 ftree& 5.0000& 1.5000&  120& 1.1000&0.7746&          tree&   house plant&     palm tree& 0.7746& 0.1193& 0.0977 \\
 ftree& 5.0000& 1.5000&  120& 1.2500&0.7717&          tree&          leaf&          bush& 0.7717& 0.1099& 0.0929 \\
 ftree& 5.0000& 1.2500&  120& 1.2500&0.7491&          tree&        grapes&          leaf& 0.7491& 0.0374& 0.0353 \\
 ftree& 5.0000& 1.5000&   60& 1.2500&0.7107&          tree&          leaf&        carrot& 0.7107& 0.0639& 0.0593 \\
pytree& 1.0000& 1.5000&   90& 1.0000&0.7085&          tree&        grapes&      broccoli& 0.7085& 0.1120& 0.0742 \\
pytree& 5.0000& 1.5000&   90& 1.0000&0.6269&          tree&           map&        grapes& 0.6269& 0.3218& 0.0164 \\
 ftree& 5.0000& 1.5000&   60& 1.1000&0.6203&          tree&           map&        grapes& 0.6203& 0.2278& 0.0899 \\
 ftree& 5.0000& 2.0000&  120& 1.0000&0.6152&          tree&     palm tree&   house plant& 0.6152& 0.3708& 0.0093 \\
pytree& 0.5000& 2.0000&   90& 1.0000&0.5329&          tree&   house plant&     palm tree& 0.5329& 0.1134& 0.0593 \\
 ftree& 5.0000& 1.5000&   90& 0.9000&0.4676&          tree&          vase&       giraffe& 0.4676& 0.1497& 0.1120 \\
 ftree& 5.0000& 2.0000&   90& 0.9000&0.4184&          tree&     hurricane&     palm tree& 0.4184& 0.2104& 0.1333 \\
 ftree& 5.0000& 1.2500&  120& 1.1000&0.3207&      broccoli&          tree&        garden& 0.3712& 0.3207& 0.0910 \\
pytree& 1.0000& 2.0000&   90& 1.0000&0.3184&          tree&        flower&        garden& 0.3184& 0.2753& 0.1662 \\
 ftree& 5.0000& 2.0000&  120& 1.1000&0.2475&     palm tree&          tree&          leaf& 0.6613& 0.2475& 0.0761 \\
 ftree& 5.0000& 1.5000&  150& 1.1000&0.2430&     palm tree&          tree&   house plant& 0.7533& 0.2430& 0.0036 \\
 ftree& 5.0000& 2.0000&  120& 0.9000&0.2387&     palm tree&          tree&   house plant& 0.7376& 0.2387& 0.0200 \\
 ftree& 5.0000& 1.2500&  150& 1.2500&0.2055&     palm tree&          tree&   house plant& 0.7930& 0.2055& 0.0006 \\
 ftree& 5.0000& 2.0000&  120& 1.2500&0.2023&     palm tree&          leaf&          tree& 0.5469& 0.2351& 0.2023 \\
pytree&10.0000& 1.5000&   90& 1.0000&0.1436&           map&          tree&     hurricane& 0.4329& 0.1436& 0.1246 \\
pytree& 0.2000& 1.5000&   90& 1.0000&0.1381&roller coaster&          tree&        garden& 0.7478& 0.1381& 0.0334 \\
 ftree& 5.0000& 2.0000&   60& 1.2500&0.1245&          leaf&          tree&     palm tree& 0.7466& 0.1245& 0.1019 \\
pytree& 2.0000& 1.0000&   90& 1.0000&0.1030&        bowtie&       hexagon&     boomerang& 0.3392& 0.2240& 0.1279 \\
pytree& 2.0000& 5.0000&   90& 1.0000&0.0952&          leaf&   string bean&          tree& 0.4812& 0.2999& 0.0952 \\
 ftree& 5.0000& 2.0000&   90& 1.1000&0.0674&          leaf&     palm tree&          tree& 0.6264& 0.1513& 0.0674 \\
 ftree& 5.0000& 1.2500&   90& 1.0000&0.0646&      broccoli&           map&         brain& 0.3342& 0.1671& 0.1553 \\
pytree& 0.5000& 1.5000&   90& 1.0000&0.0570&      broccoli&          leaf&        flower& 0.4511& 0.1738& 0.1135 \\
 ftree& 5.0000& 1.5000&   90& 1.2500&0.0527&          leaf&        carrot&        flower& 0.5632& 0.1312& 0.0683 \\
pytree&10.0000& 2.0000&   90& 1.0000&0.0444&           map&          tree&     hurricane& 0.8606& 0.0444& 0.0228 \\
 ftree& 5.0000& 2.0000&   60& 1.1000&0.0438&          leaf&          tree&           map& 0.8871& 0.0438& 0.0297 \\
 ftree& 5.0000& 1.2500&   90& 1.1000&0.0400&      broccoli&           map&          tree& 0.6310& 0.2804& 0.0400 \\
pytree& 1.0000& 5.0000&   90& 1.0000&0.0377&      scorpion&          leaf&          tree& 0.7230& 0.0467& 0.0377 \\
 ftree& 5.0000& 1.2500&  120& 1.0000&0.0361&     palm tree&          tree&          fork& 0.8003& 0.0361& 0.0312 \\
 ftree& 5.0000& 1.5000&  150& 0.9000&0.0337&     palm tree&          tree&   house plant& 0.9659& 0.0337& 0.0002 \\
 ftree& 5.0000& 1.2500&   90& 0.9000&0.0161&         brain&           cow&         fence& 0.4656& 0.2073& 0.0826 \\
pytree& 0.1000& 2.0000&   90& 1.0000&0.0097&          rake&     palm tree&roller coaster& 0.3822& 0.2686& 0.2253 \\
pytree&10.0000& 5.0000&   90& 1.0000&0.0075&          leaf&     palm tree&         broom& 0.6861& 0.2151& 0.0291 \\
 ftree& 5.0000& 2.0000&  150& 1.2500&0.0068&     palm tree&          leaf&   house plant& 0.9484& 0.0226& 0.0088 \\
 ftree& 5.0000& 1.2500&  150& 1.1000&0.0063&     palm tree&          tree&   house plant& 0.9936& 0.0063& 0.0001 \\
 ftree& 5.0000& 1.2500&  150& 1.0000&0.0054&     palm tree&          tree&         broom& 0.9944& 0.0054& 0.0000 \\
pytree& 0.2000& 2.0000&   90& 1.0000&0.0052&roller coaster&     palm tree&         fence& 0.7792& 0.0986& 0.0331 \\
pytree& 5.0000& 5.0000&   90& 1.0000&0.0037&          leaf&          bird&        dragon& 0.8446& 0.0334& 0.0211 \\
 ftree& 5.0000& 1.2500&   90& 1.2500&0.0025&      broccoli&          leaf&           map& 0.5153& 0.3316& 0.0433 \\
pytree& 0.2000& 1.0000&   90& 1.0000&0.0022&       bicycle&       octagon&     boomerang& 0.6009& 0.1402& 0.0714 \\
 ftree& 5.0000& 2.0000&  150& 0.9000&0.0019&     palm tree&          tree&   house plant& 0.9964& 0.0019& 0.0018 \\
 ftree& 5.0000& 1.5000&  150& 1.0000&0.0015&     palm tree&          tree&   house plant& 0.9983& 0.0015& 0.0002 \\
 ftree& 5.0000& 1.0000&  150& 1.0000&0.0012&     palm tree&     parachute&          rake& 0.4800& 0.3600& 0.0465 \\
 ftree& 5.0000& 2.0000&   90& 1.2500&0.0009&          leaf&     palm tree&          bush& 0.9604& 0.0261& 0.0042 \\
pytree& 1.0000& 1.0000&   90& 1.0000&0.0008&       hexagon&       octagon&        zigzag& 0.2076& 0.2073& 0.1122 \\
 ftree& 5.0000& 1.2500&   60& 1.0000&0.0008&       hexagon&           map&         brain& 0.9427& 0.0366& 0.0116 \\
 ftree& 5.0000& 1.2500&   60& 0.9000&0.0006&         brain&     boomerang&           map& 0.3458& 0.2721& 0.2014 \\
 ftree& 5.0000& 1.2500&  150& 0.9000&0.0006&     palm tree&          vase&         broom& 0.9873& 0.0099& 0.0008 \\
 ftree& 5.0000& 1.0000&  150& 0.9000&0.0005&     palm tree&     parachute&          rake& 0.4050& 0.3542& 0.1138 \\
 ftree& 5.0000& 2.0000&  150& 1.1000&0.0004&     palm tree&   house plant&          tree& 0.9978& 0.0014& 0.0004 \\
 ftree& 5.0000& 1.2500&  120& 0.9000&0.0004&     palm tree&         brain&           leg& 0.8486& 0.0693& 0.0208 \\
pytree& 0.1000& 1.0000&   90& 1.0000&0.0004&       bicycle&  eiffel tower&      triangle& 0.3852& 0.1549& 0.1367 \\
pytree& 1.0000&10.0000&   90& 1.0000&0.0004&        zigzag&   garden hose&   string bean& 0.3320& 0.1536& 0.1004 \\
 ftree& 5.0000& 1.0000&  150& 1.2500&0.0003&     parachute&     palm tree&          rake& 0.4385& 0.2488& 0.1410 \\
 ftree& 5.0000& 1.0000&  150& 1.1000&0.0003&     parachute&     palm tree&         brain& 0.8565& 0.0800& 0.0158 \\
 ftree& 5.0000& 1.2500&   60& 1.1000&0.0002&           map&       hexagon&          leaf& 0.9558& 0.0325& 0.0035 \\
pytree& 2.0000&10.0000&   90& 1.0000&0.0002&        zigzag&     boomerang&   garden hose& 0.1834& 0.1501& 0.1260 \\
pytree& 0.1000& 1.5000&   90& 1.0000&0.0002&roller coaster&          tree&        bridge& 0.9991& 0.0002& 0.0002 \\
 ftree& 5.0000& 1.2500&   60& 1.2500&0.0002&          leaf&        parrot&           map& 0.9440& 0.0198& 0.0086 \\
 ftree& 5.0000& 2.0000&  150& 1.0000&0.0002&     palm tree&   house plant&          tree& 0.9995& 0.0003& 0.0002 \\
 ftree& 5.0000& 1.0000&   90& 1.1000&0.0002&         brain&          leaf&           map& 0.6582& 0.2208& 0.0577 \\
 ftree& 5.0000& 1.0000&   90& 1.2500&0.0001&         brain&          leaf&          pear& 0.9778& 0.0163& 0.0024 \\
pytree& 0.5000& 1.0000&   90& 1.0000&0.0001&       octagon&        zigzag&     boomerang& 0.3738& 0.1488& 0.1268 \\
\hline
\end{tabular}}
}
\end{table}

Somewhat against our subjective expectations, the highest classification association with the natural trees received the fractal trees based on right-angled branching triangle and (mostly) in agreement to the da Vinci branching rule of preserved intersection. For all 6 out of 7 top trees with \textit{tree} class support standing out around or above 95\% da Vinci rule was preserved ($v=1$) and most of them (5/7) showed right branching triangle and with branch imbalance shared between $b=1.5$ (3/7) to $b=2$ (4/7). Interestingly, the $2^{nd}$ and the $4^{th}$ highest-ranked tree are actually parametrically the same, although with the imbalance of $b=2$ the random flipping could lead to quite different outcomes in height and width including a significant 1-sided leaning effect possibility. 
We also noted that two of the Pythagorean trees in the top set had much smaller elongation of $e=2$, which could mean the fixed elongation of $e=5$ applied earlier to the generalized fractal trees could have been a bit overshot. To investigate these two aspects a bit more and in attempt to further improve the Googlenet-perceived realism of our fractal tree, we regenerated more trees this time enforcing (rather than verifying) the da Vinci rule ($v=1$), but relaxing elongation in the range of $e=\{2,3,4,5,6\}$, zooming in the angles around the right angle: $\alpha=\{70\degree,80\degree,90\degree,100\degree,110\degree\}$ and zooming in around the imbalance of $b=2$, i.e. $b=\{1.5,1.75,2,2.25,2.5\}$. We have also repeated the tree generation for the same parameters 5 times to investigate the random variability of classification assessment and average them to simply get more accurate results and conclusions. Moreover, to explore the consistency of deep CNN predictions in the presence of all other classes we have also retrained 2 other more complex CNN-networks called ResNet50 \cite{ResNet50_1}, \cite{ResNet50_2} and InceptionV3 \cite{Inception3_1}, \cite{Inception3_2} that are considered more powerful and tend to produce better results than Googlenet.

This way, a set of $5\cdot 5^3=625$ fractal trees have been generated and classified by the above-mentioned 3 CNN models, producing 1875 classification results. First off we have inspected the overall classification accuracy for the whole DomainNet dataset and separately for the \textit{tree} class individually for all 3 models as shown in table \ref{tab:cnn_classification_results}.

\begin{table}
\centering
\caption{Training and validation accuracy for DomainNet real categories classification: overall and separately for the \textit{tree} class using transfer-trained CNN deep learning models}
\label{tab:cnn_classification_results}
 \begin{tabular}{r|l|l|l|l}
\hline
\textbf{\textit{}} & \makecell[l]{training \\accuracy} & \makecell[l]{validation \\accuracy} & \makecell[l]{\textit{tree} class\\training \\accuracy} & \makecell[l]{\textit{tree} class \\validation \\accuracy} \\\hline
  GoogleNet&  0.9168&    0.7814&  0.8626&    0.7631 \\
   ResNet50&  0.9687&    0.8023&  0.9439&    0.7864 \\
InceptionV3&  0.9726&    0.8145&  0.9530&    0.7987 \\\hline
\end{tabular}
\end{table}

These results confirmed slightly improved performance of the added CNN models, although it was achieved at a significant computational cost. Although the validation accuracy for all classes and the $tree$ class separately improve somewhat in tandem, the training accuracy for added networks is much higher, that suggests they might have been over-trained and some more refined fine-tuning could result in better validation accuracy, although during training the validation performance continued to rise till the last epoch.  

\begin{figure*}[hbt!]
    \centering
    \subfigure[Accuracy=f(e,b,*,1)]{\includegraphics[width=0.32\hsize]{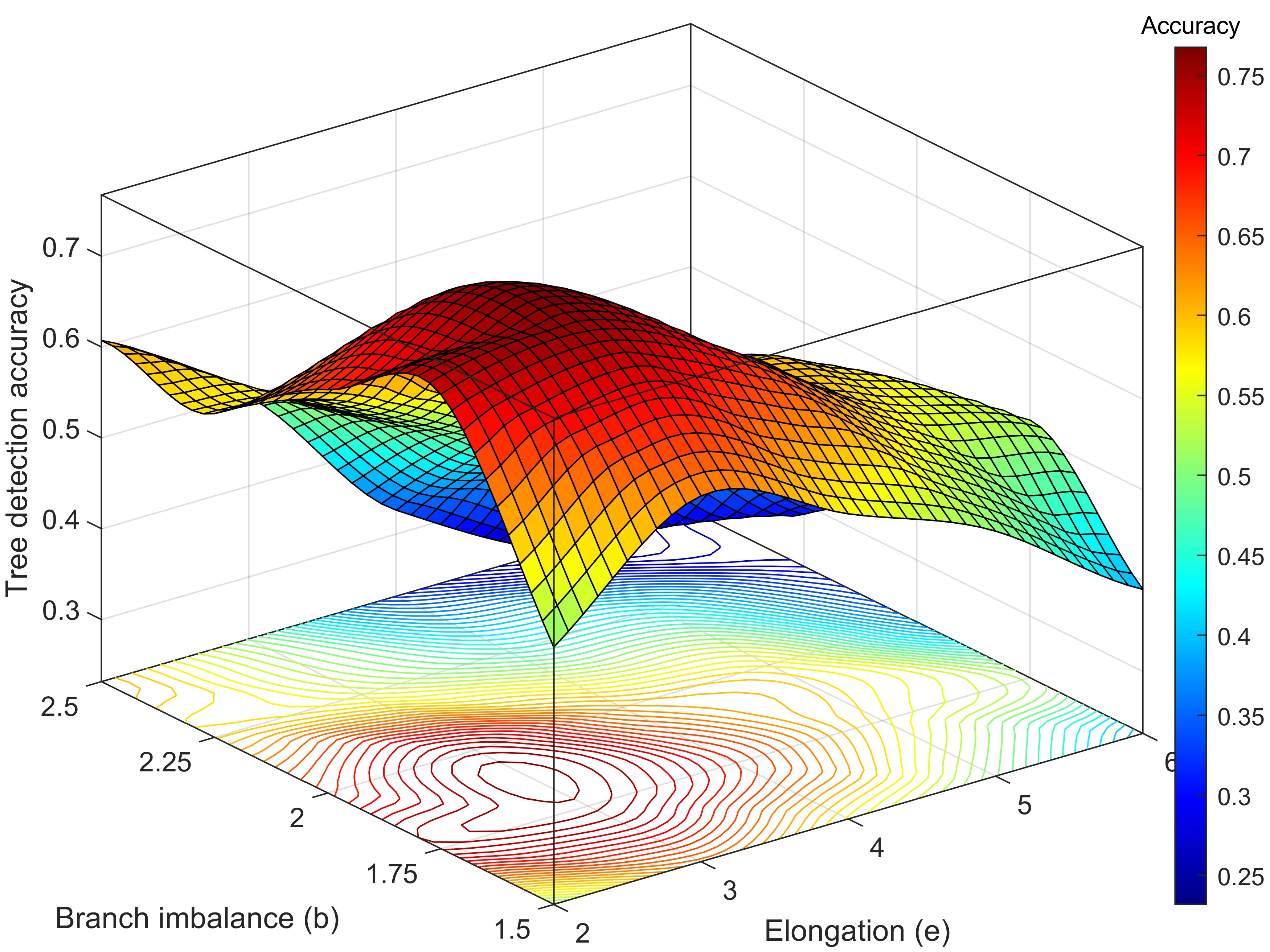}}\label{fig:acc_f_eb}
    \subfigure[Accuracy=f(e,*,$\alpha$,1)]{\includegraphics[width=0.32\hsize]{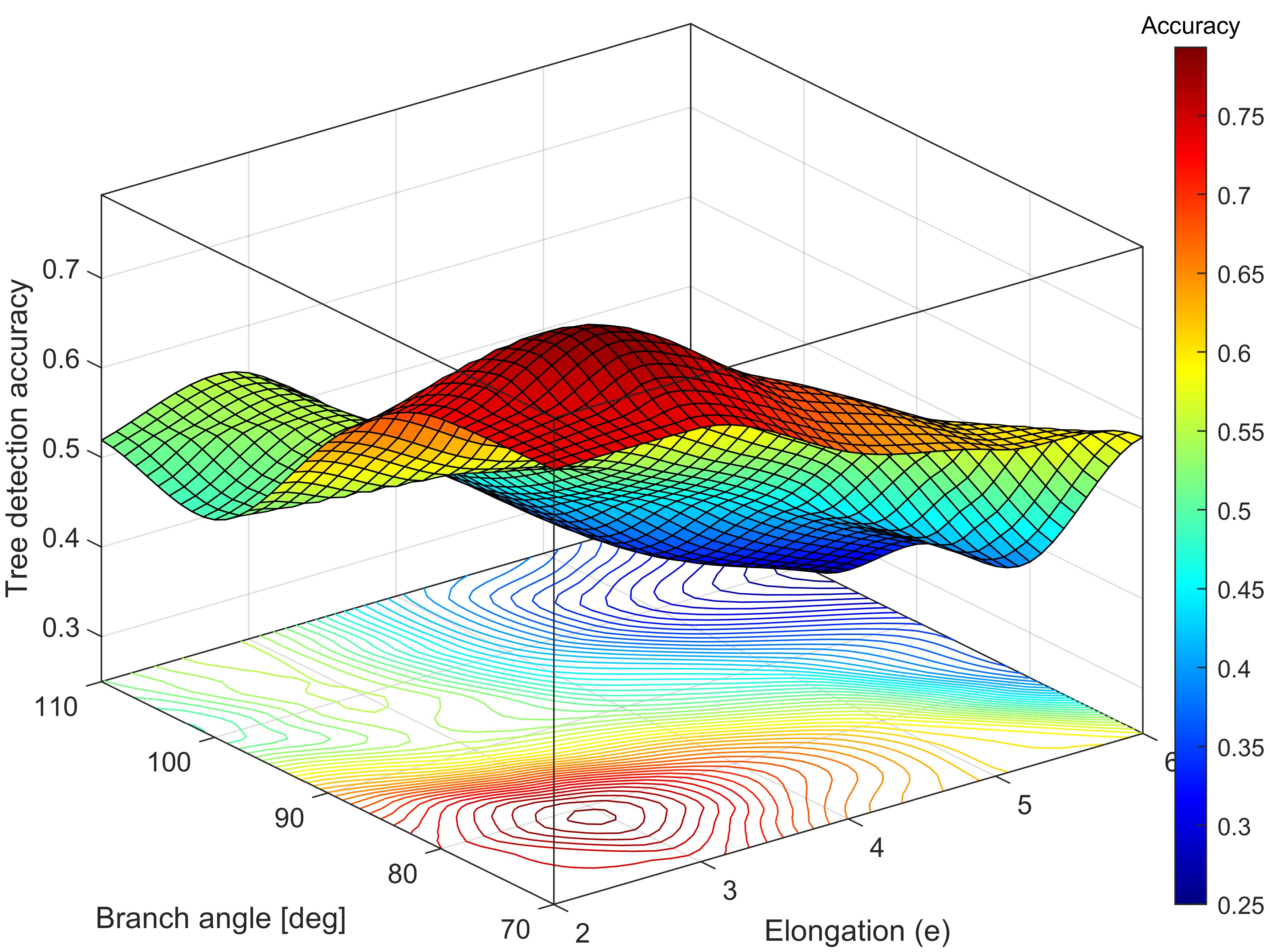}}\label{fig:acc_f_ea}
    \subfigure[Accuracy=f(*,b,$\alpha$,1)]{\includegraphics[width=0.32\hsize]{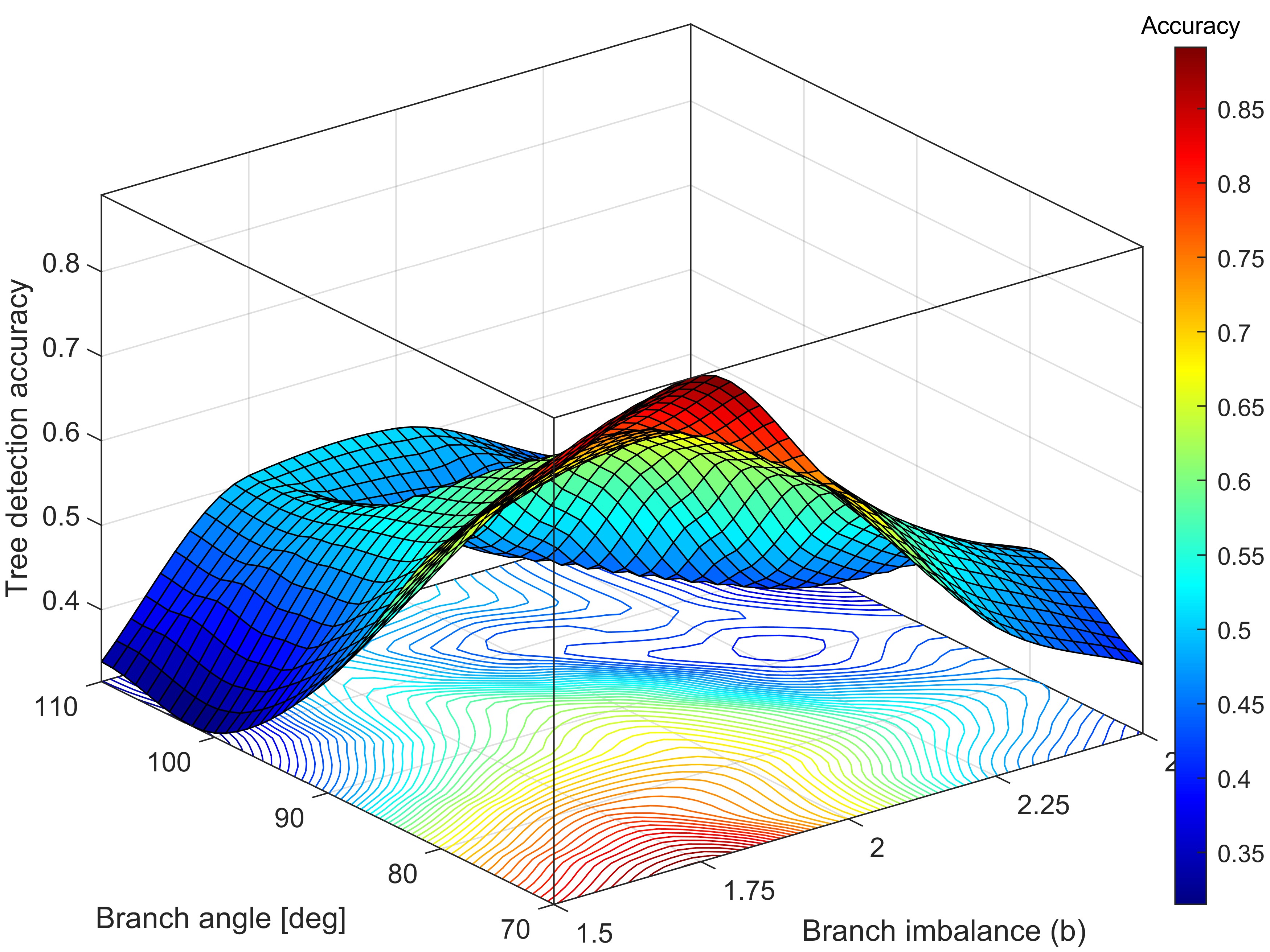}}\label{fig:acc_f_ba}
    \caption{Average accuracies of \textit{tree} classification for various pairs of parameters aggregated over 5x regenerated fractal trees of $T(e,b,\alpha,1)$ family with preserved Da Vinci rule ($v=1$) classified by 3 different CNN transfer-trained networks: GoogleNet, ResNet50 and InceptionV3}  
    \label{fig:param_surface_plots}
\end{figure*}

Going back to the classification results of our fractal trees, we have first aggregated the accuracy for various combinations of pairs of the e,b,$\alpha$ parameters to obtain parametric performance surface plots depicted in Figure \ref{fig:param_surface_plots}.

\begin{figure*}[hbt!]
    \centering
    \includegraphics[width=1\hsize]{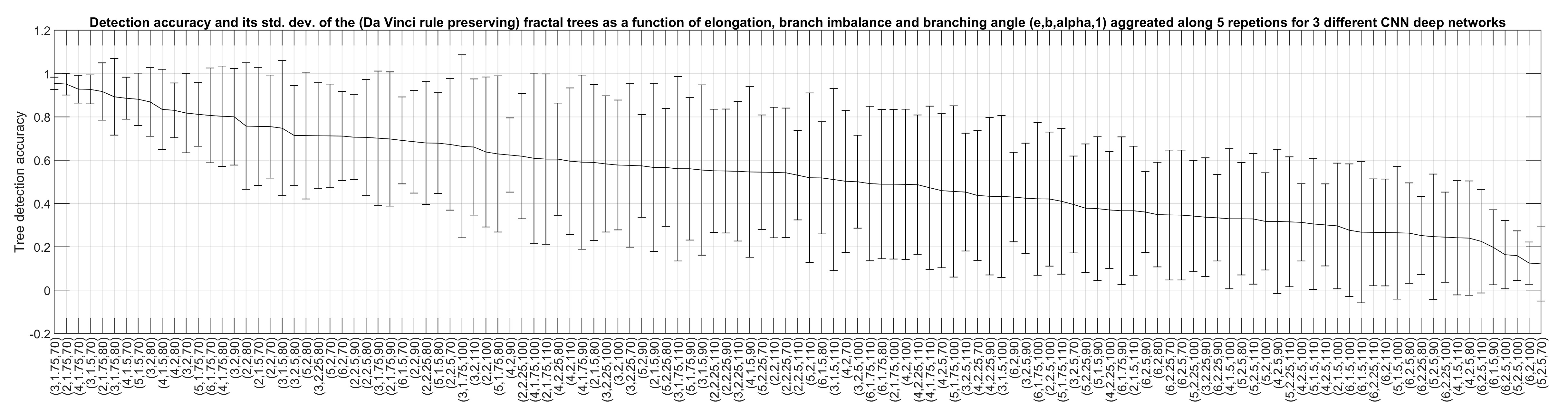}
    \caption{Average sorted accuracies and standard deviations of \textit{tree} classification for all parametric combinations from $T(e,b,\alpha,1)$ family} 
    \label{fig:acc_f_eba}
\end{figure*}

These results point very clearly at highest tree classification accuracy for elongation $e=3$, branch imbalance of $b=1.75:2$ and angle between the branches of $\alpha=70:80\degree$. More granular parametric performance results are shown in Figure \ref{fig:acc_f_eba}. Here the tree classification accuracy is aggregated for combinations of all 3 variable parameters $(e,b,\alpha,1)$ and presented descendingly along with standard deviations obtained along only 15 values (5 repetitions with 3 CNN networks). Very clearly the best results in terms of both accuracy and consistency are observed for branching angle of $\alpha=70\degree$ and average elongation of $e=3$ although scattering from 2 to 4. The optimal branch imbalanced is also confirmed to be $b=1.75$ although $b=1.5$ is also strongly present around the top (3/8).

\begin{figure}[hbt!]
    \centering
    \includegraphics[width=1\hsize]{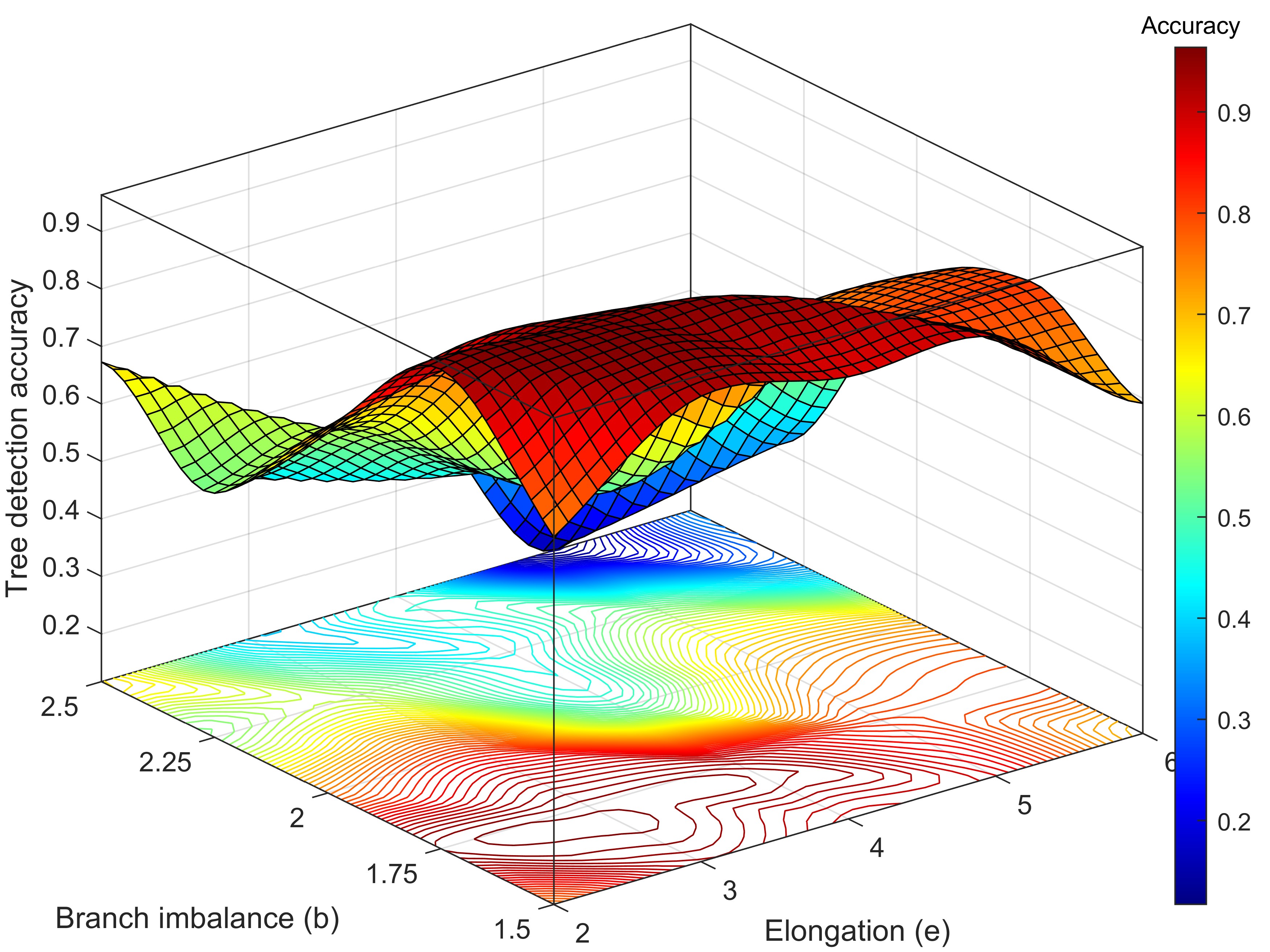}
    \caption{Interpolated surface plot of average accuracies of \textit{tree} classification for various combinations of (e,b) parameter pairs from $T(e,b,70,1)$ family} 
    \label{fig:acc_f_eb70}
\end{figure}

Here the tree classification accuracy is aggregated for combinations of all 3 variable parameters $(e,b,\alpha,1)$ and presented descendingly along with standard deviations obtained along only 15 values (5 repetitions with 3 CNN networks). Very clearly the best results in terms of both the accuracy and its consistency are observed for the branching angle of $\alpha=70\degree$ and the average elongation of $e=3$, although scattering from 2 to 4. The optimal branch imbalanced is also confirmed to be $b=1.75$ although $b=1.5$ is also strongly present around the top (3/8). Based on the feedback from Figure \ref{fig:acc_f_eba}, we have regenerated the parametric surface plot of tree detection accuracy yet this time further constrained by the branching angle fixed at $\alpha=70\degree$ as shown in Figure \ref{fig:acc_f_eb70}.


\begin{figure}
    \centering
    \includegraphics[width=\hsize,height=19.4cm]{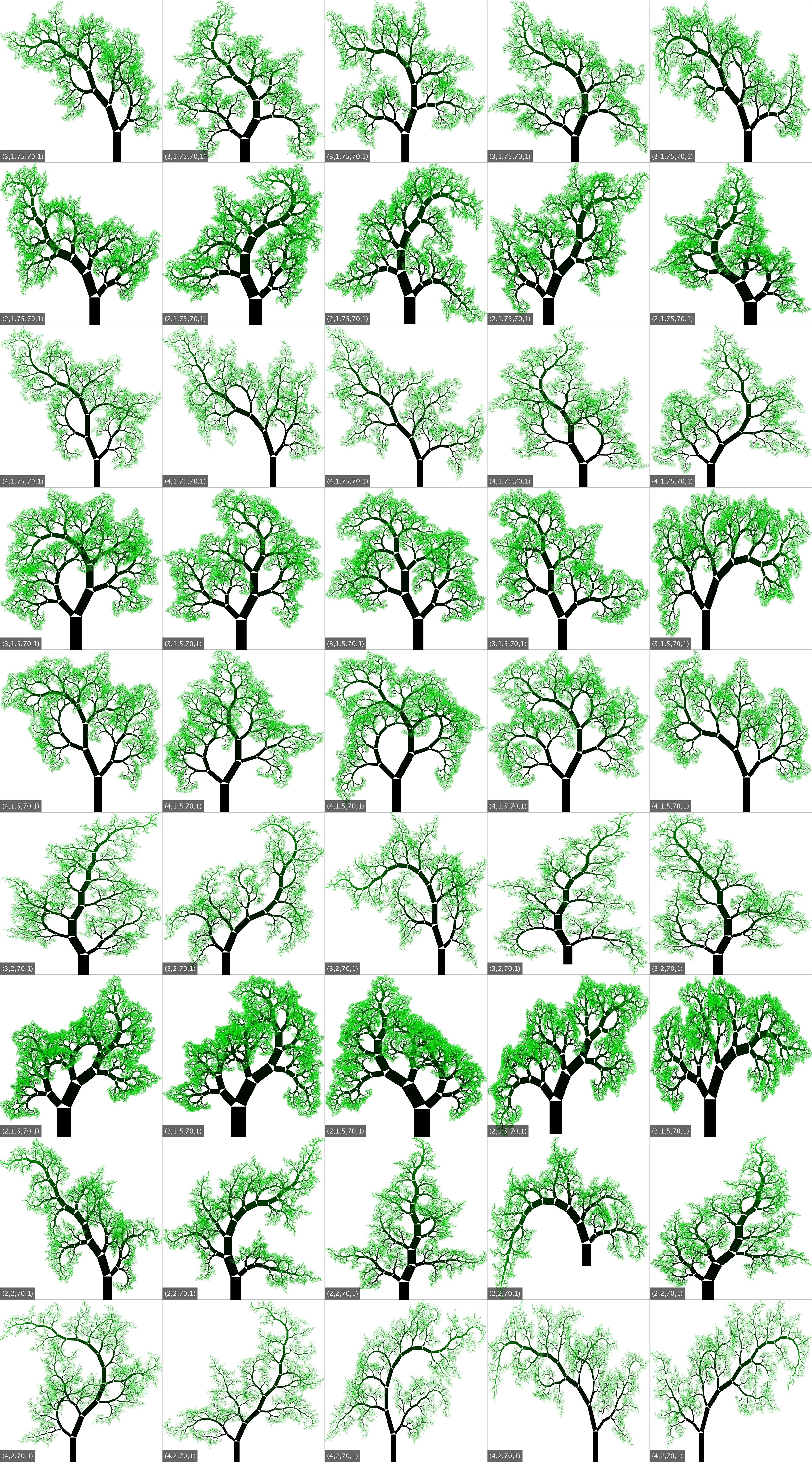}
    \caption{Tree variants in the optimal parametric plateau: $T(2:4,1.5:2,70\degree,1)$.} 
    \label{fig:ftrees_top}
\end{figure}

A very clear plateau above 90\% accuracy emerges for the elongation range of $e=(2,4.5)$ and the imbalance range of $b=(1.5,1.9)$. The fractal trees corresponding to this optimal parametric corner are illustrated in Figure \ref{fig:ftrees_top}. This time these trees are resized to the exactly same square format that the CNN models \textit{see} or rather require as input for both training and prediction.     

Further empirical investigations pointed at the optimal tree parameters of $T\!=\!(3.15,1.62,67\degree,1)$ consistently producing trees detectable by CNNs at the accuracy of above 99\%. Interestingly, the optimal branch imbalance coincides nearly perfectly with the golden ratio of $g=(1+\sqrt{5})/2\approx=1.618$, which uniquely is equal to the ratio of the sum to the larger number, while the optimal elongation is not far from the doubled golden ratio. Whether these are coincidental observations is yet to be investigated in a subsequent study.

Before concluding the analysis we took the optimal tree we identified as $T(2g,g,67\degree,1)$ and dived into the code again to see the actual internal transformation parameters for it. The scales for the children branches initially were: $\{s_l=0.7323,s_r=0.4526\}$; yet, after re-normalization to enforce da Vinci rule, they grew up to $\{s_l=0.8507,s_r=0.5257\}$ in both cases maintaining golden ratio of $s_l/s_r=g=1.618$. What is interesting, is that assuming both da Vinci rule and golden ratio of the children branches' are obeyed, the actual scales no longer depend on the branching angle $\alpha$ but become entirely determined by the golden ratio:

\begin{equation}
    s_l^2+s_r^2=1,\; \frac{s_l}{s_r}=g \; \implies \; s_r=\frac{1}{\sqrt{g^2+1}}, \;s_l=gs_r
\end{equation}

The rotation angles, however turned out to be $\{\gamma=24.62\degree,\beta=42.38\}$, corresponding to the left and right child-branch rotations, respectively. These rotation angles incidentally remain at the ratio of $\beta/\gamma=1.72$, which is again quite close to the golden ratio $g$. What if the ratio of rotation angles is also forced to be exactly golden: $\beta/\gamma=g=1.618$?. In such case the only actual controllable parameter remaining would be the branching angle $\alpha$ and since $\alpha=\beta+\gamma$ the choice of $\alpha$ will uniquely determine both rotation angles that preserve the golden ratio:
\begin{equation}
    \frac{\beta}{\gamma}=\frac{\beta+\gamma}{\beta}=\frac{\alpha}{\beta}=g\;\;  \implies \;\; \beta=\frac{\alpha}{g} \;\; \gamma=\frac{\alpha}{g^2}
\end{equation}

\begin{figure*}[ht!]
    \centering
    \includegraphics[width=1\hsize]{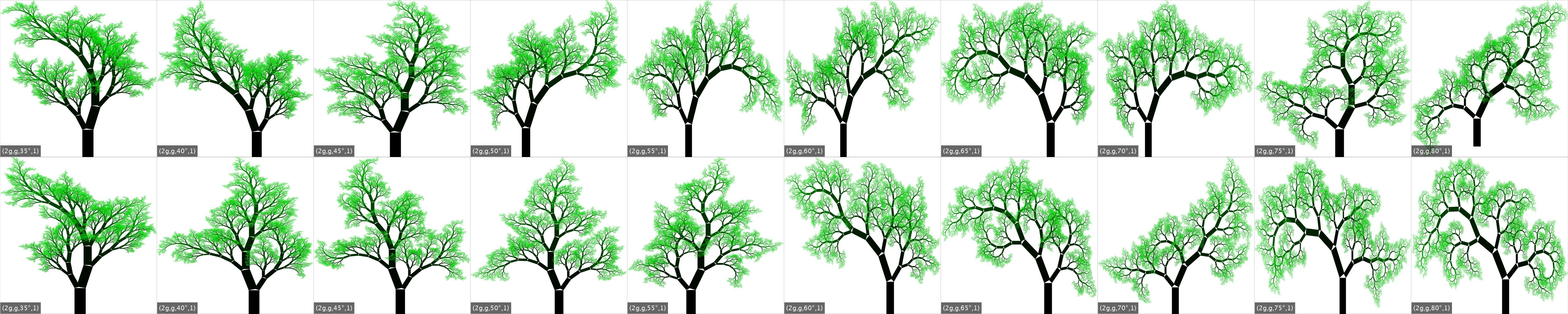}
    \caption{Comparison of the most realistic fractal trees (top row) with \textit{golden} trees (bottom row) for which the rotation angles of child branches were additionally constrained to preserve the golden ratio, along incremental branching angle from $\alpha=35\degree$ to $\alpha=80\degree$.} 
    \label{fig:fvg_trees}
\end{figure*}

Since the golden ratio is fixed, the branching angle $\alpha$ would remain the only controllable parameter or degree of freedom that would vary the tree.
Figure \ref{fig:fvg_trees} shows the comparison of the sample fractal trees for incremental $\alpha$: $T(2g,g,35:80\degree,1)$ and the corresponding \textit{golden} trees (bottom row) with rotation angles additionally constrained by the golden ratio: $\beta/\gamma=g$. There is no significant visual differences, although when testing their \textit{realism} with our trained CNN models the average \textit{tree} class accuracy for $20\times$ regenerated trees for a reasonable set of $\alpha=(45:5:80)$ point at better results for the golden trees especially for sharper branching angles, as shown in Table \ref{tab:gtree_perf}. 

\begin{table}[ht] \centering
\caption{CNN-assessed \textit{realism} of fractal trees $T(2g,g,\alpha,1)$ compared to the \textit{golden} trees with rotation angles locked at golden ratio}
\label{tab:gtree_perf}
\renewcommand{\arraystretch}{0.6} {\setlength{\extrarowheight}{4pt}
 \begin{tabular}{|r|l|l|l|l|l|l|}
\hline
\rowcolor{gray!20} Tree$\backslash \alpha$ &$45\degree$ &$50\degree$ &$55\degree$ &$60\degree$ &$65\degree$ &$70\degree$ \\ [1ex] 
\hline
ftree & 0.4323& 0.6139& 0.8950& 0.9490& 0.9820& 0.9026 \\
gold  & 0.5054& 0.7284& 0.9108& 0.9535& 0.9841& 0.8948 \\
\hline
\end{tabular}}\end{table}

Concluding, the fractal trees that appear to be the most realistic according to representative CNN deep learning models are experimentally determined to belong to the $T(2g,g,67\degree,1)$ family, with (after some preliminary testing) further improvements possible when the golden ratio constraints are further imposed upon the rotation angles of the children branches. 
A shockingly compact fractal prescription on just a few lines of the recursive code takes a rectangle of height $2g\times$ greater than the width, scales, shifts and rotates recursively into pairs of branches maintaining golden ratio, preserving da Vinci's rule and going apart by an angle of about $\alpha=67\degree$, produces a branching structures that when colored with depth-aligned linear color-map from black to green, look indistinguishably close to the natural trees.        

\section{Conclusion}\label{sec:Conclusions}
We have set off on a journey to try to reconstruct the natural tree side look, with the simplest possible fractal model that could be literally written in a few lines of a recursively coded algorithm. To achieve this goal we have modified the Pythagorean tree fractal algorithm and paired it with a flexible parameterization structure that takes rectangular trunk and appends it recursively with pairs of similar down-scaled branches at fully controlled parameters: branch elongation ($e$), branch imbalance ($b$), branching angle ($\alpha$) and the \emph{da Vinci factor} accounting for thinner or thicker intersection of the children branches compared to the parent branch. These parameters were mapped into fixed pre-computed constants used throughout the efficient matrix-represented recursive operations of scaling, translation and rotation to rapidly generate population of deep fractal trees with flip-randomised forking and branches colored with a linear black-to-green color-map. We have explored the parametric space of such remarkably short recipe and identified a parametric plateau that tends to produce very naturally-looking trees. 
We have quantitatively evaluated the realism or the \textit{natural look} of generated fractal trees by the \textit{tree} class classification score of the convolutional neural network (CNN) predictors competitively cross-trained on the hundreds of natural tree images along with hundreds of thousands of other images from 345 classes. Interestingly, the most naturally looking fractal tree family identified as $T(2g,g,67\degree,1)$ is composed of rectangles elongated by twice the golden ratio ($2g$), branches imbalanced by exactly the $g$ and spread by about $67\degree$ apart, while obeying both the da Vinci and Grigoriev's intersection or surface area preservation rules ($v=1$). The trees generated from the $T(2g,g,67\degree,1)$ family are consistently classified by the CNN predictor with natural tree label and its support to the \textit{tree} class exceeding $0.95$. Despite very simplistic representation and rather neglected concept of leaves,  the generated fractal trees look very natural to the human eye. Further work in this space will concentrate on improvements in weight-aware spatial balancing of the trees, especially relevant when attempting to build stable tree models in 3D. We also aim to build upon this work to develop the capability to fractal-reconstruction of the specific tree species aimed at generating artificial training examples to build a powerful ML image-based tree species predictor and compare it against much more computationally expensive generative AI models. The authors are also enthusiastic to explore a logarithmic fractal tree representation paired with Fourier transformation analysis that can complement the research of Grigoriev et al. \cite{Grigoriev2022} and perhaps extend it beyond the deciduous trees.

\bibliographystyle{tocplain}   


\begin{tocauthors}
\begin{tocinfo}[ruta]
    Dymitr Ruta\\
    Chief Researcher\\
    EBTIC, Khalifa University\\
    Abu Dhabi, United Arab Emirates\\
    dymitr.ruta \tocat{} ku \tocdot{} ac \tocdot{} ae
\end{tocinfo}

\begin{tocinfo}[mio]
    Corrado Mio\\
    Senior Researcher\\
    EBTIC, Khalifa University\\
    Abu Dhabi, United Arab Emirates\\
    corrado.mio \tocat{} ku \tocdot{} ac \tocdot{} ae
\end{tocinfo}
\begin{tocinfo}[damiani]
    Ernesto Damiani\\
    Professor, Dean of Computing and Mathematical Sciences\\
    Director of Center for Secure Cyber- Physical Systems\\
    Khalifa University\\
    Abu Dhabi, United Arab Emirates\\
    ernesto.damiani \tocat{} ku \tocdot{} ac \tocdot{} ae
\end{tocinfo}
\end{tocauthors}

\begin{tocaboutauthors}

\begin{tocabout}[ruta]  

\textsc{Dymitr Ruta} is the Chief Scientist at the Emirates ICT Innovation Center (EBTIC) at Khalifa University (KU), leading the Big Data Analytics and Machine Learning Group. He earned his PhD in Statistical Pattern Recognition from the University of the West of Scotland (UWS) in 2003. His research interests include machine learning (ML), artificial intelligence (AI), and mathematical modeling and simulation. Before joining EBTIC, Dr. Ruta was a Lecturer at UWS, a Senior Researcher at British Telecom Lab (UK), and a Senior Researcher at Boronia Capital Hedge Fund (Australia). He is an accomplished researcher with over 100 publications, more than 6 patents, and 20+ international awards. Dr. Ruta leads AI/ML-driven research and innovation projects, delivering state-of-the-art predictive services focused on data monetization for large industrial and government partners. He has a proven track record of generating multimillion-dollar value in the finance, telecom, aerospace, education, and health industries.
\end{tocabout}

\begin{tocabout}[mio]
\textsc{Corrado Mio} is Senior researcher at the Emirates ICT Innovation Center (EBTIC) at Khalifa University (KU). He earned his PhD in Computer Science from Università degli studi di Milano (Italy) in 2020. His research interests include Machine Learning (ML), NLP, Massive Parallel and Distributed Algorithms, and Programming Languages. Before joining EBTIC, Dr Mio worked in the R\&D department in several companies as software developer/architect and team leader where he has acquired experience in software development and implementation of complex software architectures. He has strong experience in all main programming languages, paradigms and platforms. He has also strong experience in parallel and distributed programming, real-time and fault-tolerant applications, back-end and front-end applications. He has also worked in Joint Research Centre (Ispra, Italy) where he implemented Ray Tracing algorithms (on multiple bands) to realize solar radiation models.
\end{tocabout}

\begin{tocabout}[damiani]
\textsc{Ernesto Damiani} ... short bio ....
\end{tocabout}

\end{tocaboutauthors}

\end{document}